\PassOptionsToPackage{noend}{algorithm2e}
\documentclass[pmlr,twocolumn,10pt]{jmlr} 

\usepackage{booktabs}
\usepackage{siunitx}
\usepackage{bbm}
\usepackage[bf]{caption}
\usepackage[defaultcolor=red]{changes}
\usepackage[export]{adjustbox}
\newtheorem{innercustomthm}{Theorem}
\newenvironment{customthm}[1]
  {\renewcommand\theinnercustomthm{#1}\innercustomthm}
  {\endinnercustomthm}
\usepackage[all]{nowidow}
\usepackage[switch]{lineno}

\newcommand{\equal}[1]{{\hypersetup{linkcolor=black}\thanks{#1}}}

\definechangesauthor[name={Aishwarya}, color=orange]{AM}
\definechangesauthor[name={Matt}, color=orange]{MJ}
\colorlet{Changes@Color}{orange}

\newcommand{\argmax}{\mathop{\mathrm{arg\,max}}}

\theorembodyfont{\upshape}
\theoremheaderfont{\scshape}
\theorempostheader{:}
\theoremsep{\newline}

\jmlrvolume{LEAVE UNSET}
\jmlryear{2024}
\jmlrsubmitted{LEAVE UNSET}
\jmlrpublished{2024}
\jmlrworkshop{Conference on Health, Inference, and Learning (CHIL) 2024} 

\title[Adaptive Interventions with User-Defined Goals for Health Behavior Change]{Adaptive Interventions with User-Defined Goals for Health Behavior Change}
\author{%
\Name{Aishwarya Mandyam}\equal{These authors contributed equally} \Email{am2@stanford.edu}\\
\addr Stanford University
\AND
\Name{Matthew J{\"o}rke}\footnotemark[1] \Email{joerke@stanford.edu}\\
\addr Stanford University
\AND
\Name{William Denton}
\Email{wdenton@stanford.edu}\\
\addr Stanford University
\AND
\Name{Barbara E. Engelhardt} \Email{barbarae@stanford.edu}\\
\addr Gladstone Institutes, Stanford University
\AND
\Name{Emma Brunskill} \Email{ebrun@cs.stanford.edu}\\
\addr Stanford University
}

\begin{document}

\maketitle
\begin{abstract}
Promoting healthy lifestyle behaviors remains a major public health concern, particularly due to their crucial role in preventing chronic conditions such as cancer, heart disease, and type 2 diabetes. Mobile health applications present a promising avenue for low-cost, scalable health behavior change promotion. Researchers are increasingly exploring adaptive algorithms that personalize interventions to each person's unique context. 
However, in empirical studies, mobile health applications often suffer from small effect sizes and low adherence rates, particularly in comparison to human coaching. 
Tailoring advice to a person's unique goals, preferences, and life circumstances
is a critical component of health coaching that has been underutilized in adaptive algorithms for mobile health interventions. To address this, we introduce a new Thompson sampling algorithm that can accommodate personalized reward functions (i.e., goals, preferences, and constraints), while also leveraging data sharing across individuals to more quickly be able to provide effective recommendations. We prove that our modification incurs only a constant penalty on cumulative regret while preserving the sample complexity benefits of data sharing. We present empirical results on synthetic and semi-synthetic physical activity simulators, where in the latter we conducted an online survey to solicit preference data relating to physical activity, which we use to construct realistic reward models that leverages historical data from another study. Our algorithm achieves substantial performance improvements compared to baselines that do not share data or do not optimize for individualized rewards.

\end{abstract}
\paragraph*{Data and Code Availability}
Our simulators depend on publicly available data from \cite{milkman2021megastudies}\footnote{\url{https://osf.io/9av87/?view_only=8bb9282111c24f81a19c2237e7d7eba3}} and data we collected in an online study. All code and de-identified data are available at \url{https://github.com/StanfordAI4HI/adaptive-interventions-with-goals}. 
\paragraph*{Institutional Review Board (IRB)}
Our online study was approved by Stanford's Institutional Review Board (\#46241).

\section{Introduction}
\label{sec:intro}
Chronic conditions are a leading cause of mortality, 
accounting for 70\% of deaths~\citep{watson2022chronic} and a substantial portion of healthcare expenditures~\citep{buttorff2017multiple} in the United States. Importantly, many chronic conditions, including cancer, heart disease, type 2 diabetes, and obesity, are preventable through the adoption of healthy lifestyle behaviors~\citep{watson2022chronic}, such as nutritious eating, regular physical activity, and avoiding tobacco and alcohol consumption. Health coaching is a popular and effective intervention for promoting health behavior change~\citep{olsen2010health, wolever2013systematic}. While in-person health coaching is effective, it is also expensive, inaccessible to many patient populations, and does not scale to global need~\citep{bickmore2011reusable, mitchell2021automated}.

Researchers are increasingly exploring mobile health (mHealth) applications as a low-cost, scalable, and accessible approach to motivate health behavior change~\citep{domin2021smartphone, hicks2023leveraging}. 
Within machine learning and statistics, there is a burgeoning interest in personalizing health behavior change interventions by applying adaptive experimentation or reinforcement learning algorithms to automatically discover which interventions work best for different individuals across diverse contexts~\citep{mintz2019nonstationary, heartsteps, baek2023policy, ruggeri_benzerga_verra_folke_2023}. 
Despite the potential for low-cost, personalized, contextually-tailored interventions to promote positive health outcomes, mHealth interventions are known to suffer from small effect sizes~\citep{yang2019comparative} and low adherence~\citep{yang2020factors},
particularly in comparison to human health coaches~\citep{mcewan2016effectiveness}. 

A key component of effective health coaching is goal setting~\citep{olsen2010health, wolever2013systematic, epton2017unique}.
Effective goals are both (1) conscious and specific, and (2) sufficiently difficult. For example, an ineffective goal is vague (\textit{``I want to get more exercise''}), whereas an effective goal is specific and sufficiently challenging (\textit{``I want to go on a brisk walk for 30 minutes each day''}). 
Goal-setting theory highlights that effective goals focus attention towards goal-related activities and lead to greater effort and persistence~\citep{locke2002building}. 
While goal setting strategies have been explored in prior mHealth and self-tracking tools~\citep{goals_goalsetting}, support for personalized goals is largely absent from algorithms for adaptive interventions. Instead, prior work typically optimizes for some shared, measurable outcome (e.g., step count), implicitly assuming that each person wants to maximize this quantity (e.g., more steps is always better). Not only does this approach neglect the positive psychological effects of goal setting on long-term motivation, it also ignores the diversity of people's goals~\citep{epton2017unique}. Moreover, when individuals receive feedback on misaligned or overly ambitious goals, it can lead to abandonment or habituation \citep{locke2002building, peng2021habit}.

In addition, an important role of a health coach is to personalize their advice to a client's unique goals, habits, and life circumstances such that they are empowered to adopt healthier lifestyle behaviors~\citep{rutjes2019beyond, wolever2013systematic, olsen2010health, olsen2014health}.
For example, people may have different time constraints (e.g., work or family obligations), physical abilities (e.g., preferring low-impact activities), or access to resources (e.g., living near a park), all of which influence which interventions are appropriate.
The combination of goal-setting and personalization to preferences and constraints is missing from existing adaptive interventions in mHealth literature, even though these are both important aspects for effective behavior change. 

Perhaps surprisingly, we show that both goal setting and constraints on interventions can be addressed in a unified framework. We present a new algorithm that optimizes for user-specific reward functions, where each of these reward functions can encode user-specific preferences or constraints over outcomes and interventions.

The contributions of this work to the existing literature for adaptive interventions are as follows: 
\begin{enumerate}
    \item We present a new Thompson sampling algorithm for linear contextual bandits~\citep{agrawal2013thompson} that optimizes individualized reward functions (i.e., goals, preferences, and constraints) while maintaining the ability to share information across individuals. We formalize user goals as a Lipschitz continuous functions of a shared outcome variable (e.g., step count) that has common structure across individuals. 
    \item We provide a bound on the cumulative regret of our approach. Our regret bound matches comparable bounds for linear contextual bandits that do not consider personalized reward functions. 
    \item We apply our approach to a modified version of an existing simulator for physical activity~\citep{liao2016sample} and a semi-synthetic simulator for gym attendance based on data from a large-scale study by~\cite{milkman2021megastudies} and an online preference study that we conducted on Prolific. Our experiments highlight that it is possible to simultaneously optimize for user-specific goals and respect user preferences while sharing data across a cohort of users. Notably, our algorithm outperforms policies trained separately for each user and policies that optimize for a misspecified reward that ignores personal goals. 
\end{enumerate}
\section{Related Work}
\label{sec:relatedwork}
The potential of adaptive experimentation in mobile health lies in machine learning algorithms being able to recommend the most effective intervention when it is most likely to prompt action. 
Such algorithms can make use of contextual information, measured via passive sensing (e.g., mobile phones, wearables) or self-reported data, to tailor interventions to each person's unique state~\citep{jitai}. An adaptive and personalized approach to intervention delivery is appropriate for health behavior change, where people differ substantially across many factors that are important for effective personalization, such as their habits, goals, abilities, environments, life circumstances, and more~\citep{hicks2023leveraging}. 

Prior work has explored a number of algorithmic approaches for personalizing interventions in mobile health~\citep{mintz2019nonstationary,liao2019personalized, heartsteps, baek2023policy, ruggeri_benzerga_verra_folke_2023}. Several studies have formulated learning optimal policies for sending notifications as a bandit problem~\citep{Rabbi2015MyBehaviorAP} that can consider contextual variables describing a user's state~\citep{poptherapy, diabetes_activity}. 
In empirical studies, contextual algorithms can improve physical activity, but adherence to notifications decreases over time~\citep{heartsteps}. 
In an ideal setting, a full Markov decision process (MDP) would be used to model the long-term effect of a recommended action on a user's state. However, bandit approximations are common in the literature as MDPs typically require too much data to be feasible in the settings of interest. 
One approach more accurately models users by assuming the reward is non-stationary and proposes a variant of the upper confidence bound (UCB) algorithm~\citep{mintz2019nonstationary}. As a step towards using a full MDP, another method estimates one step of policy iteration within a high-dimensional MDP and learns optimal policies using a model-free algorithm~\citep{baek2023policy}. In an empirical study, the resulting policy performed just as well as standard procedures, but required only half the budget. 

However, all prior algorithmic methods optimize for some shared outcome, such as maximizing step count~\citep{liao2019personalized, klasnja2015microrandomized, diabetes_activity}, calorie loss~\citep{Rabbi2015MyBehaviorAP}, or stress reduction~\citep{poptherapy}. 
This outcome is chosen by researchers, implicitly assuming that all participants want to optimize this outcome. 
Most algorithms also do not consider personalized preferences or constraints over the types of interventions or their delivery.
Meanwhile, health coaching programs advocate for a client-centric and goal-oriented approach that respects client autonomy and aims to empower clients to achieve their own health-related goals~\citep{olsen2010health}. 
A client-centric and goal-oriented reframing of adaptive experimentation in mobile health actively involves participants, helping them discover which interventions work best for them in achieving their own health goals. 
Additionally, prior work in personal informatics and self-tracking has designed systems that allow people to set their own goals~\citep{goals_goalsetting}, but these approaches have yet to be explored in the literature on adaptive interventions.

One simple approach to optimizing over individualized goals is to train an independent policy for each person, using their own goal as a reward function. However, this approach does not leverage the shared structure of data across individuals and thus requires more samples to learn an optimal intervention assignment rule.
Prior work has explored sharing data in settings where multiple agents concurrently explore the same environment~\citep{pmlr-v28-silver13, Guo_Brunskill_2015, Pazis_Parr_2016,dimakopoulou2018coordinated} or where one agent learns across multiple related task environments~\citep{deshmukh2017multitask, kveton2021metathompson, hong2022hierarchical}. Other approaches have improved sample efficiency using collaborative filtering to cluster users ~\citep{gentile2014online, gentile2017contextdependent}.  Our setting is distinct in that we propose a hybrid environment in which outcomes can be pooled across individuals while reward functions cannot.
Existing work on optimizing individual preferences
focuses on efficiently learning complex, unknown preferences for a single user~\citep{interactive_multiobjective} or adapting to an individual's preference changes over time~\citep{garivier2008upperconfidence, Hariri2015AdaptingTU, Wu_2018, luo2019efficient}. 
In contrast, we assume that preferences are known
for each user and the objective is to quickly maximize individualized rewards across all users. 
\section{Preliminaries}
\label{sec:preliminaries}
We formalize our setting as a Bayesian contextual bandit environment. We assume there are a set of $P$ individuals who, at each time point, are eligible to receive an intervention.  At each time $t$, each individual  $i \in [P]$ is in a state (also known as context) $s_{i,t}$ that is sampled independently from a known distribution $s_{i, t} \sim \rho$.
There are a set of possible actions (also known as interventions) $a \in [A]$ that can be recommended to each individual. A decision policy is a mapping from the state of an individual $s_{i,t}$ to a recommended action $a_{i, t} \in [A]$. For example, if an individual has not yet met their daily step goal, the policy might send a notification with a motivational message. After choosing action $a_{i,t}$ in state $s_{i,t}$, a reward $r_{i,t} \in \mathbb{R}$ is observed. We will further describe the reward function and its structure below (Eq.~\ref{eq:reward}). Crucially, this reward is not known in advance. This procedure repeats for $t \in [T]$ rounds. The goal of this algorithm is to learn, through observing the contexts, actions, and rewards, a policy that maximizes the cumulative sum of rewards for a single individual.

A critical part of such adaptive algorithms is representing the context of an individual. Here we 
let $\boldsymbol{\phi}_{i, t} \triangleq \boldsymbol{\phi}(s_{i,t}, a_{i,t}) \in \mathbb{R}^d$ be a known state-action featurization. For example, $\boldsymbol{\phi}_{i, t}$ may contain demographic information about user $i$ (e.g., age, gender, pre-existing conditions), action-specific information (e.g., indicator variables for a given treatment or treatment interaction terms), and other contextual information (e.g., time, day of the week, location, number of previous interventions). 
Let $\mathbf{y}_{i,t} \triangleq \mathbf{y}(s_{i,t}, a_{i,t}) \in \mathbb{R}^M$ be a vector containing $M$ outcomes, which are each measurable quantities of interest about a user. For example, $\mathbf{y}_{i,t}$ can contain measurements such as step count, sleep duration, heart rate, or blood pressure. 
In our setting, 
we assume each entry $m$ in $\mathbf{y}_{i,t}$ follows a linear model,\footnote{While we assume linearity for Theorem \ref{thm:bayesianregret_main}, Algorithm~\ref{alg:ts_aug} is amenable to non-linear extensions so long as $\mathbf{y}$ admits an efficiently computable posterior given the history.}
\begin{equation}
    [\mathbf{y}_{i,t}]_m = \boldsymbol{\phi}_{i,t}^\top \boldsymbol{\theta}_m + \varepsilon
    \label{eq:y-simple}
\end{equation} 
where $\varepsilon$ is independent Gaussian noise.
Note that this model is shared across all users, and each $\boldsymbol{\theta}_m$ is specific to outcome $m$. In our Bayesian setting, we assume that each $\boldsymbol{\theta}_m$ is drawn from a known prior $p(\boldsymbol{\theta})$. Our assumption of shared structure in $\mathbf{y}_{i,t}$ allows our algorithm to pool information across users for improved sample complexity.

In prior work, it is common to use contextual bandits to optimize directly for an 
outcome $\mathbf{y}_{i,t}$ (e.g., maximizing step count).
Instead, we construct an objective that allows us to optimize for each user's unique goals while accommodating their preferences over how they would like to achieve that goal. 
We formalize our reward function as a weighted sum of $K$ user-specific utilities. Specifically, upon taking action $a_{i, t}$ in state $s_{i,t}$, the environment reveals reward
\begin{align}
    r_{i,t}(s_{i, t}, a_{i,t}) &= 
    \sum_{k=1}^K w_{i,t,k} \cdot U_{i,t,k}([\mathbf{y}_{i,t}]_{m_k}, \boldsymbol{\phi}_{i, t}).
    \label{eq:reward}
\end{align}
We define a utility function $U_{i,t,k} : \mathbb{R}^M \times \mathbb{R}^d \rightarrow \mathbb{R}$ to be a mapping from outcomes $\mathbf{y}_{i,t}$ and contexts $\boldsymbol{\phi}_{i,t}$ to a scalar that represents the degree to which a goal is met or preferences are satisfied.
We assume that each utility function $U_{i,t,k}$ depends only on the $m_k$th entry of $\mathbf{y}_{i,t}$ and is $L$-Lipschitz in $[\mathbf{y}_{i,t}]_{m_k}$, i.e., $\forall y, |U(y, \phi) - U(y', \phi)| \leq L|y - y'|$. 
A utility $U_{i,t,k}$ can encode preferences over outcomes (e.g., walking 10,000 steps per day), preferences over intervention types (e.g., preferring prompts for reflection over nudges to exercise), or preferences over intervention delivery (e.g., penalizing excessive notifications), all of which may be context-specific (e.g., preferring no interventions during weekdays or a higher step count target on weekends). 

Our reward function is composed of multiple utility functions to reflect that each user may have many, possibly conflicting, goals and preferences.
To reduce this multi-objective problem to a scalar reward signal, each utility function is associated with a known weight $w_{i,t,k}$ satisfying $\sum_{k=1}^K w_{i,t,k} = 1$. 
These weights are either fixed by the algorithm designer or elicited from the user to reflect their preferences over the relative importance of each utility function. 

We allow all $w_{i,t,k}$ and $U_{i,t,k}$ to change for each user $i$ at any time $t$, which enables users to update their goals and preferences over time as their priorities change. 
This implies that $r_{i,t}$ is non-stationary, which violates standard bandit assumptions and would make sample-efficient learning challenging in full generality. 
However, since each $r_{i,t}$ is a known and deterministic function of $\mathbf{y}_{i,t}$, which itself is a stationary function that is shared across users, 
it is possible to efficiently learn each $\boldsymbol{\theta}_m$ despite optimizing for personalized, time-dependent reward.
\begin{algorithm2e*}
\caption{Multi-Objective Multi-User Thompson Sampling}
\raggedright
\label{alg:ts_aug}
\DontPrintSemicolon
\SetAlgoNoLine
\Indentp{0.4em}
\KwIn{number of participants $P$, number of timesteps $T$, number of outcomes $M$, prior $p(\boldsymbol{\theta}) \sim \mathcal{N}(0, \lambda^{-1}I)$, user-specific utilities $U_{i,t,k}$ and weights $w_{i,t,k}$, $\forall i\in[P], k\in[K]$}
\vspace{0.33em}
For all $m \in [M]$, let $p_m = p(\boldsymbol{\theta})$, $\mu_m = \mathbf{0}$, $b_m = \mathbf{0}$, and $\Sigma = \lambda I$ \; 
\For{$t=1,\dots,T$}{
\hspace{-10pt}
    \For{$i=1,\dots,P$}{ 
        \hspace{-20pt}
        Sample parameters from posterior $\tilde{\boldsymbol{\theta}}_m \sim p_m$ and compute outcomes:
        $[\tilde{\mathbf{y}}(s_{i,t}, a)]_m = \boldsymbol{\phi}(s_{i,t}, a)^\top \tilde{\boldsymbol{\theta}}_m$\;  
        \hspace{-20pt}
        Choose action to maximize weighted utility using known $U_{i,t,k}$ and  $w_{i,t,k}$:
        $a_{i,t} = \argmax_{a}\sum_{k=1}^K w_{i,t,k} U_{i,t,k}\left([\tilde{\mathbf{y}}(s_{i,t}, a)]_{m_k}, \boldsymbol{\phi}(s_{i,t}, a)\right)$ \;
        \hspace{-20pt}
        Observe the true outcome $\mathbf{y}_{i, t}$ and update the posterior:
        $\Sigma \leftarrow \Sigma + \boldsymbol{\phi}_{i,t}\boldsymbol{\phi}_{i,t}^\top, 
        \quad \forall m \in [M] : b_m \leftarrow b_m + \boldsymbol{\phi}_{i,t} [\mathbf{y}_{i, t}]_m , 
        \quad \mu_m \leftarrow \Sigma^{-1}b_m,
        \quad p_{m} \leftarrow \mathcal{N}(\mu_m, \Sigma^{-1})$
    }
}
\vspace{0.33em}
\end{algorithm2e*}

\section{Methods}
\label{sec:methods}
We present a new Thompson sampling (TS) algorithm for linear contextual bandits~\citep{agrawal2013thompson} that can optimize for user-specific goals while sharing data across users. We first present our algorithm and then prove a cumulative regret bound.

\paragraph{Algorithm}
TS is a popular contextual bandit algorithm that selects actions based on the posterior probability they are optimal, $a_t \sim P(a = a^\star | s_t, \mathcal{H}_t)$, where $\mathcal{H}_t$ represents the history of states, actions, and rewards. It uses a 1-sample estimator of the optimal action by sampling rewards $\tilde{r}_t(a) \sim P(r_t | s_t, a, \mathcal{H}_t)$ for each action and choosing $a_t = \argmax \tilde{r}_t(a)$.

We propose a modification to TS that propagates sampled outcomes through user-specific reward functions.
At each time $t$, the algorithm samples potential outcomes $\tilde{\mathbf{y}}(s_{i,t}, a)$ for each action from the posterior. Rather than directly optimizing for an outcome, our algorithm propagates outcomes through the known, user-specific reward function, modeled as a weighted sum of utility functions (Eq.~\ref{eq:reward}). The algorithm then chooses the actions that optimizes the user-specific reward function and updates the posterior after observing the resulting outcome $\mathbf{y}_{i,t}$. Our procedure is described in Algorithm \ref{alg:ts_aug}. 
\paragraph{Theoretical results}
While our primary contribution is algorithmic and empirical, to provide some assurance of our algorithm's performance, we also analyze performance with respect to Bayesian cumulative regret (BR). Cumulative regret characterizes the expected difference between the maximum possible reward and the reward achieved by some policy, summed over all timesteps.  
We measure Bayesian regret as a function of $N = P \cdot T$, the total number of samples observed by the algorithm.
\begin{equation}
\scalebox{0.89}{ 
$\displaystyle
    \hspace{-0.5em}
    \text{BR}(N) = \mathbb{E}\left[ 
    \sum_{t=1}^T \sum_{i=1}^P \max_{a \in [A]} r_{i,t}(s_{i,t}, a) - r_{i,t}(s_{i,t}, a_{i,t})
    \right].
    \label{eq:bayes-regret}
$}
\end{equation}
The expectation is taken both over the prior $p(\boldsymbol{\theta})$ and the policy trajectories. 
Since we measure regret with respect to our non-stationary, user-specific reward signal $r_{i,t}$ (Eq.~\ref{eq:reward}), our notion of regret is somewhat distinct from classic stationary regret. 

In Theorem \ref{thm:bayesianregret_main}, we demonstrate that by propagating uncertainty through the known user-specific utilities, we achieve similar regret bounds to simpler contextual multi-armed bandit approaches. 
\begin{theorem}
\label{thm:bayesianregret_main}
After running Algorithm \ref{alg:ts_aug} for $N=P \cdot T$ samples, where $P$ is the number of participants in the cohort, $T$ is the time horizon, and $M$ is the number of outcomes, we achieve Bayesian cumulative regret on the order of
\begin{equation}
    \text{BR}(N) \leq O(Ld \sqrt{N \log(NM)\log(N/d)}).
\end{equation}
\end{theorem}
The proof is provided in Appendix \ref{appendix:regret} and follows from a modification of \cite{lattimore2020bandit} Theorem 36.4, which establishes an $O(d\sqrt{N \log(N) \log(N/d)}$ regret bound for standard Thompson sampling with linear contextual bandits. Our algorithm is thus able to match standard $\widetilde{O}(d\sqrt{N})$ regret bounds (up to a factor of $L$) despite optimizing for user-specific reward functions.

The computational cost of our algorithm scales linearly with the number of users and time steps. For each user $i$ at each time $t$, we sample a parameter from the Gaussian posterior, observe outcomes, and update the posterior. This has the same complexity as linear regression and for any reasonable $d$ can be run in real time on edge devices.
\section{Experiments}
\label{sec:experiments}
We now demonstrate the results of our algorithm using two simulators: (1) a synthetic simulator inspired by mobile health intervention studies such as HeartSteps~\citep{klasnja2015microrandomized, heartsteps} and (2) a semi-synthetic simulator based on a large-scale gym attendance study~\citep{milkman2021megastudies} that we merge with an online preference elicitation study that we conducted with crowdworkers. In both simulators, we show that our approach achieves lower cumulative regret than baseline methods, highlighting our algorithm's unique ability to recommend actions that allow a user to achieve their personal goals while simultaneously accommodating their preferences. 

\subsection{Step Count Simulator}
The HeartSteps study~\citep{klasnja2015microrandomized} aims to improve users' physical activity outcomes by adaptively sending notifications encouraging them to walk. We draw inspiration from prior simulators designed to model the HeartSteps study~\citep{effect_changes, yao2021power}, with additional modifications specific to our setting.

\begin{figure*}
\begin{minipage}[t]{.48\textwidth}
\vspace{0pt}
\centering 
\includegraphics[width=\linewidth]{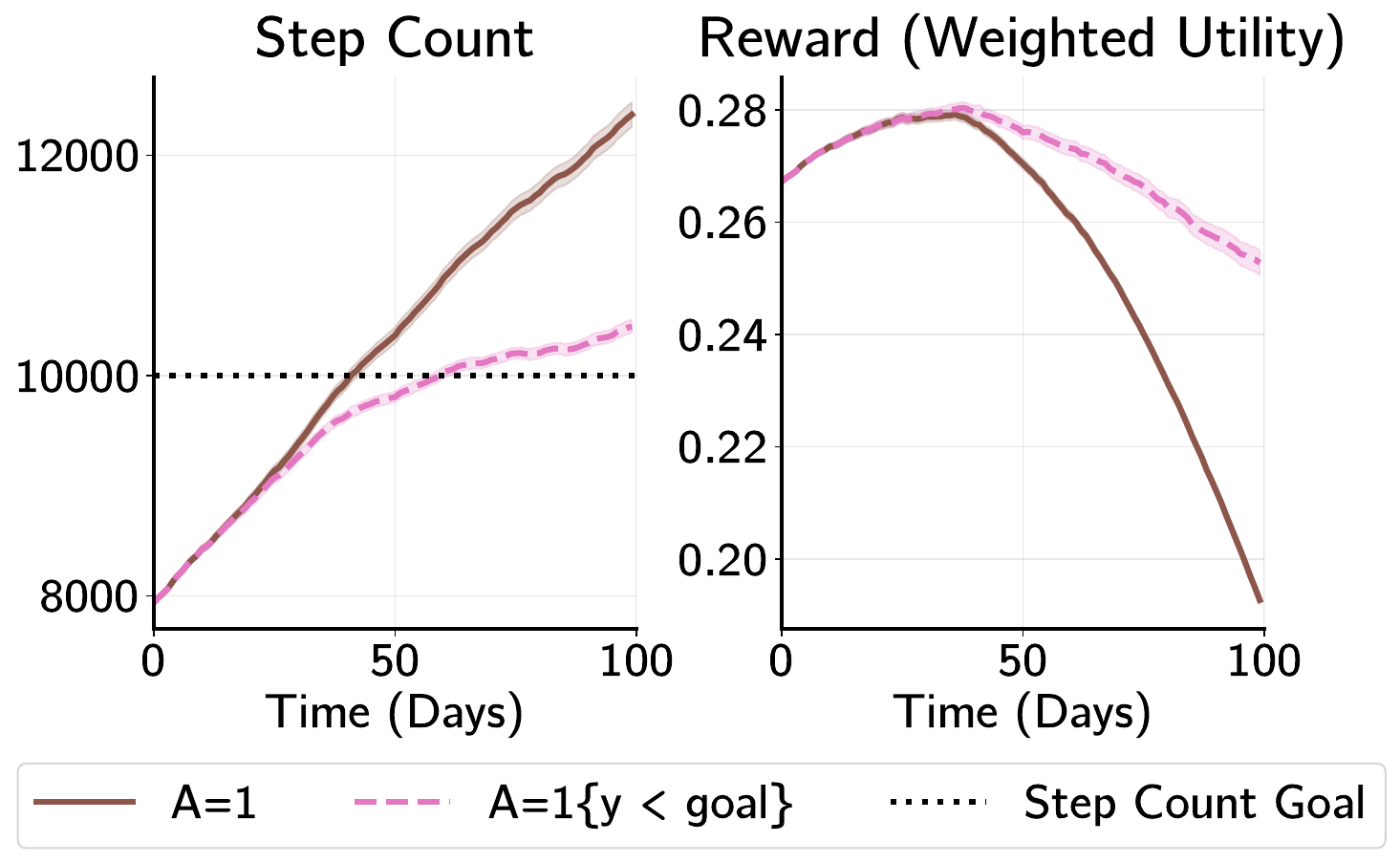}
  \caption{We compare step count $y_{i,t}$ and reward $r_{i,t}$ (Eq.~\ref{eq:reward}) for two policies: one policy ($A=1$, {\color{Chocolate4} brown}) always sends notifications and another policy ($A = \mathbbm{1}\{y < \text{goal}\}$, {\color{magenta} pink}) sends notifications only when a user's step count drops below their desired goal (dashed line). A policy that always sends a notification achieves higher step count, but lower reward due to notification burden. Shaded area is one standard error across $P=20$ participants.}
  \label{fig:heartsteps-reward-vs-utility}
\end{minipage}
\hfill
\begin{minipage}[t]{.48\textwidth}
\vspace{0pt}
\centering
\includegraphics[width=1.005\linewidth]{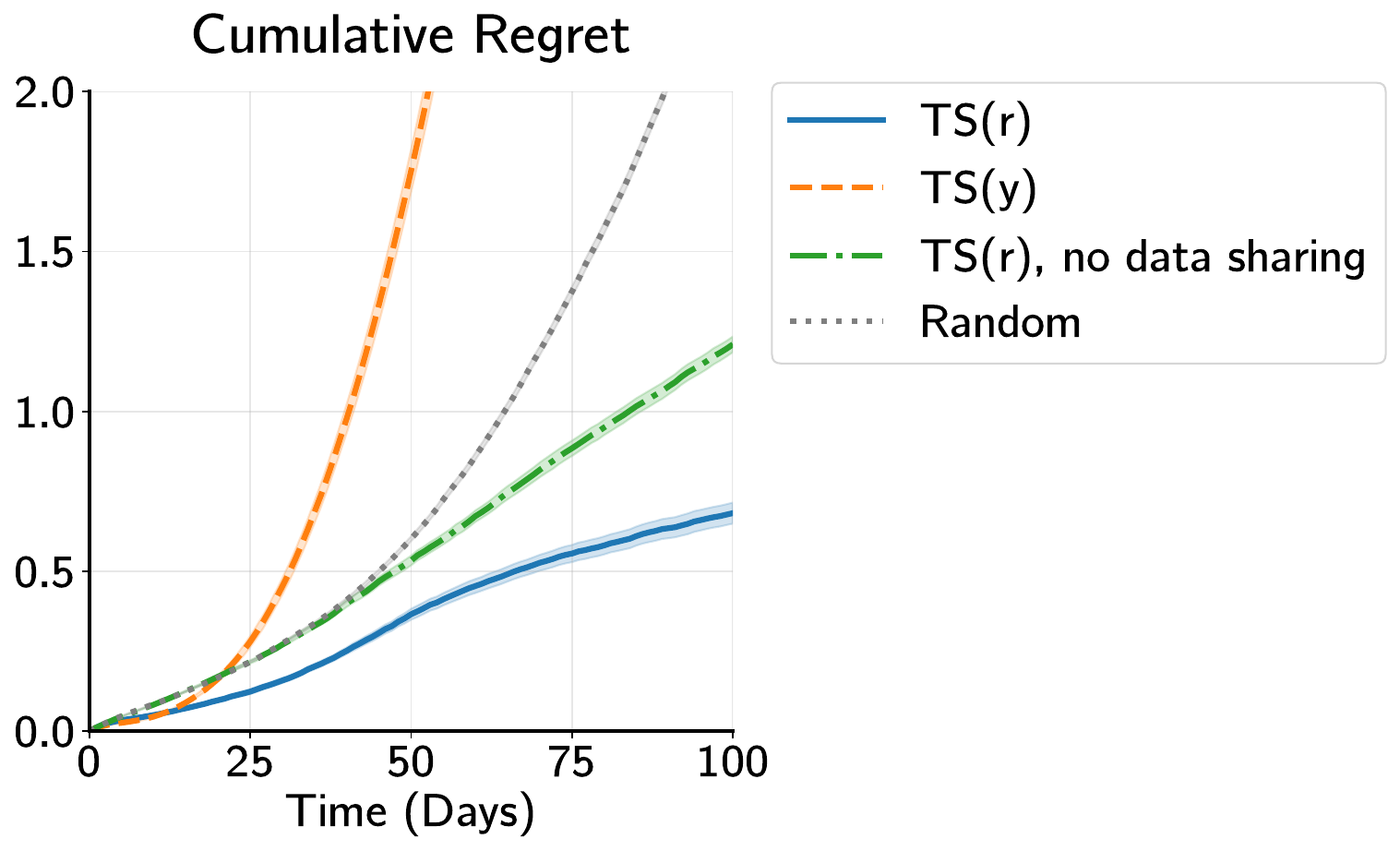}
  \caption{We compare our Algorithm \ref{alg:ts_aug} ($\textsf{TS}(r)$, {\color{blue} blue}) to TS optimizing for step count ($\textsf{TS}(y)$, {\color{orange} orange}), a random policy ({\color{gray} gray}), and independent TS policies for each user ($\textsf{TS}(r)$ no data sharing, {\color{Green4} green}). 
  We plot cumulative regret, which measures the sum of differences between the maximum possible reward and the reward achieved by a given policy at each timestep.
  Our algorithm achieves the best performance because it optimizes for the correct objective that considers notification burden and shares data across users. Shaded area is standard error across 100 trials with $P=40$ participants.}
  \label{fig:heartsteps-cum-regret}
\end{minipage}
\vspace{-1em}
\end{figure*}

First, we model treatment effect heterogeneity and non-stationarity using two clusters of users: active and inactive users. 
Empirically, studies have found that some users consistently respond positively to notifications, whereas others' response declines over time~\citep{effect_changes}. The MyHeart Counts study~\citep{SHCHERBINA2019e344}, which contains a similar setup to HeartSteps without adaptive interventions, also describes clusters of users with varying baseline activity levels and treatment responses. 
Next, we adopt an autoregressive structure to model individual variation. This allows for an intervention at timestep $t$ to impact future timesteps $t'>t$ while preserving a bandit framework, allowing us to model habit formation~\citep{hagger2019habit, peng2021habit}. Lastly, we keep track of the number of notifications that a user has received over the course of the study in order to model notification burden. \citet{notification_burden} notes that receiving too many notifications has a negative impact on a user's engagement, and we want to explicitly model the trade-off between sending notifications and increasing step count.

To construct our simulator (described in full detail in Appendix \ref{appendix:heartsteps}), 
we model one outcome $y_{i,t}$ (step count) as a function of $\boldsymbol{\phi}_{i,t}$, which contains information such as the previous day's step count $y_{i,t-1}$ and group-specific treatment effects. 
We define two utility functions: $U_{i,t,1}$, a piece-wise linear function that models step count goals, and $U_{i,t,2}$, which penalizes higher numbers of notifications. Inspired by habituation and recovery dynamics evidenced in prior work, we choose a quadratic function to represent notification burden~\citep{mintz2019nonstationary, data_driven_interpretable}. 
We model group differences by assuming that inactive users have a lower baseline activity level and treatment effect than active users, with individual variation resulting from the autoregressive structure in each population. We assume that active users place a higher weight on their step count goal and a lower weight on the notification penalty, and vice versa for inactive users. 

Under this setting, there is a trade-off between sending notifications and accumulating reward. To demonstrate this, we plot step count and reward (i.e., user-specific weighted utility) for two policies: one that always sends notifications and another that only sends notifications when a user's step count is below their goal (Figure \ref{fig:heartsteps-reward-vs-utility}). We evaluate these policies on a simulated set of 20 participants, all from the active group. 
Our results demonstrate that it is possible to consistently increase step count by always sending notifications, but that this leads to lower reward because it induces notification burden. In contrast, a policy that sends fewer notifications can still enable users to achieve their step count goal, but incurs less of a notification penalty. Intuitively, this means that a policy that does not consider preferences may perform poorly with respect to user-specific reward. 

Next, we present results comparing four algorithms:  Thompson sampling (TS) optimizing for $r_{i, t}$ ($\textsf{TS}(r)$, Algorithm \ref{alg:ts_aug}), TS optimizing for $y_{i, t}$ ($\textsf{TS}(y)$), an independent $\textsf{TS}(r)$ policy for each user that does not share data, and a random policy. While Algorithm \ref{alg:ts_aug} and Theorem \ref{thm:bayesianregret_main} assume sequential treatment assignment (i.e., the policy is updated between user $i$ and user $i+1$ at time $t$), we use a more realistic parallel treatment assignment in our experiments, updating the posterior at time $t$ only after treatment has been assigned to all users. We evaluate algorithm performance with respect to Bayesian cumulative regret (Eq.~\ref{eq:bayes-regret}), which we plot as a function of $t$ by summing instantaneous regret across users. We use a simulated population of 40 participants with 20 active and 20 inactive participants.

Our results demonstrate that TS optimizing for $r_{i, t}$ while sharing data generates the lowest cumulative regret across our horizon ($T=100$) because it correctly balances the number of notifications sent while optimizing for user-specific goals (Figure \ref{fig:heartsteps-cum-regret}). Additionally, we get a speedup in cumulative regret when we share data, indicating that sharing data allows us to learn the optimal policy for each user more quickly. Meanwhile, the random and $\textsf{TS}(y)$ policies are insensitive to notification burden and incur a quadratic notification penalty. Note that the shape of the cumulative regret curves depends heavily on the utility functions---in this example, the random and $\textsf{TS}(y)$ have superlinear cumulative regret because the utility corresponding to notification burden is a negative quadratic function.

\subsection{Gym Attendance Semi-Synthetic Simulator}
A key assumption to our method is that people have diverse preferences and goals, and that accounting for these is important in adaptive experimentation algorithms to support health behavior change. To validate these assumptions, we conducted an online study to better understand physical activity goals and preferences. 
We created a semi-synthetic simulator by combining this preference data with (separate) historical data from a large-scale study on gym attendance~\citep{milkman2021megastudies}.
This semi-synthetic simulator allows us to understand the potential benefit of our method under realistic data distributions. We believe this is a necessary precursor to conducting a field study that would require multiple years of development and substantial financial resources.

In this section, we first describe our online survey, the gym attendance dataset, and our regression-based simulator. We then evaluate our bandit algorithm in the semi-synthetic simulator.

\paragraph{Online Preference Study}

We conducted an online survey on Prolific with 220 participants to collect preferences over different physical activity interventions. 
Participants in the online study all lived in the US, spoke fluent English, and were diverse across several demographic variables and geographic location (see Appendix \ref{appendix:24h:online-study}). We excluded participants who failed attention checks or whose completion time was one or more standard deviations below the mean, leaving a total of 209 participants after exclusion. 

We found that participants varied greatly in their baseline levels of physical activity and types of physical activity they engage in. The most common activities were walking, cardio, and strength training. 51\% of participants reported engaging in less than the recommended 150 minutes per week of physical activity~\citep{who_pa}. Most participants wanted to increase their physical activity, with 76\% agreeing or strongly agreeing with the statement, \textit{``I would like to change my current levels of physical activity.''}

\begin{figure}[h]
    \vspace{0pt}
    \centering
    \includegraphics[width=1\linewidth]{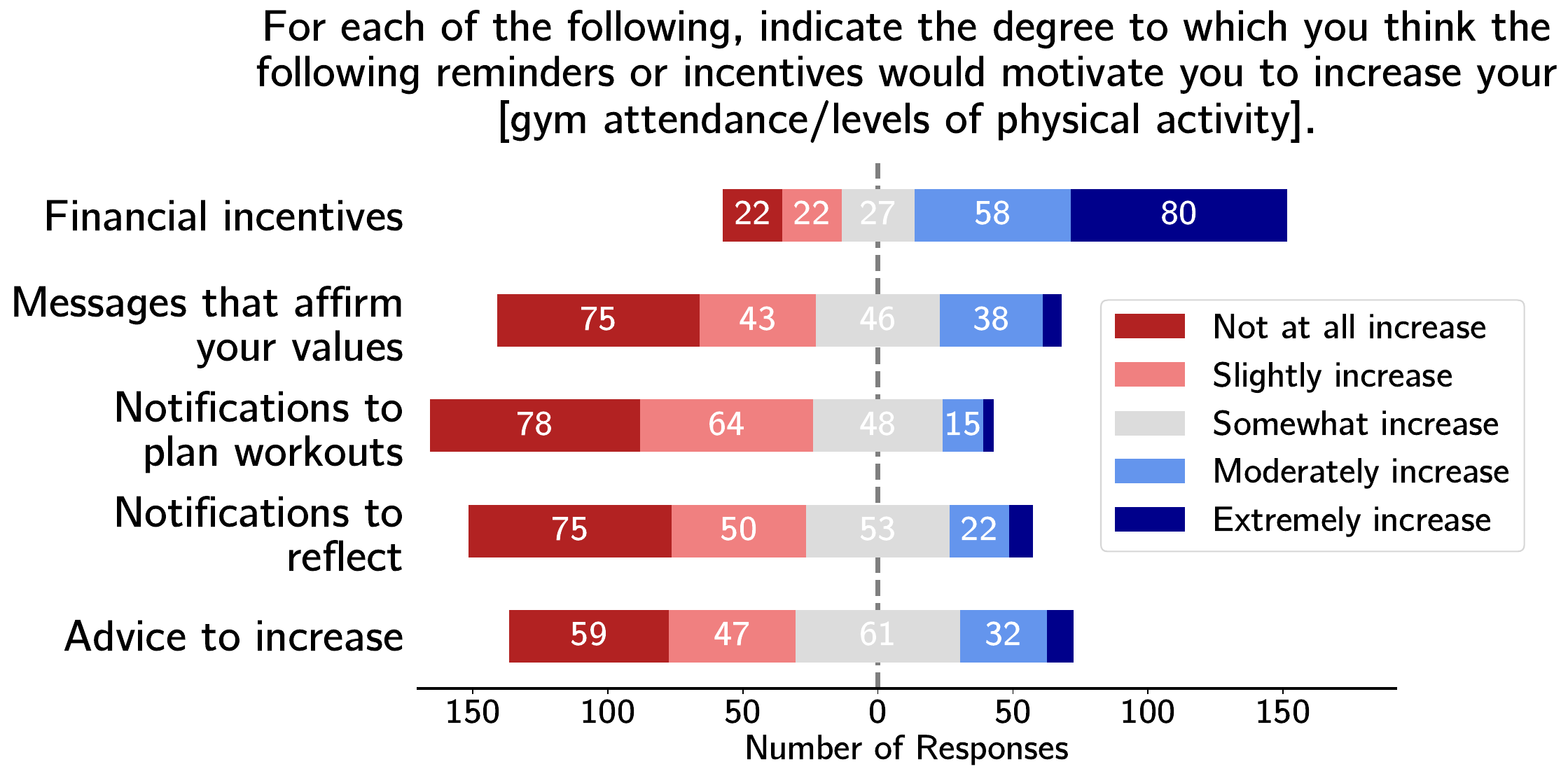}
    \caption{Participant preferences over the five intervention categories in our online survey, which we use to compute a preference vector $\boldsymbol{\alpha}_i$. While financial incentives are most popular and notification are least popular on average, participant varied widely in their individual preference ratings.}
    \label{fig:alpha-preferences}
\end{figure}
\begin{figure}[h]
    \centering
    \includegraphics[width=\linewidth]{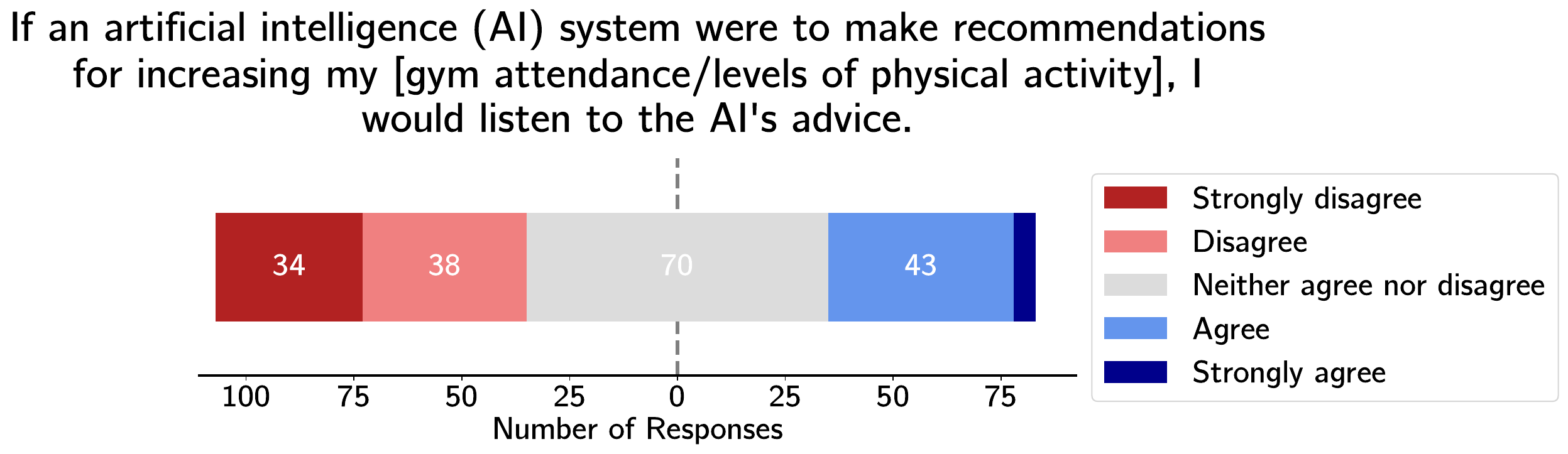}
    \caption{Participant preferences over the degree to which they would listen to an AI's advice, which we use to compute a preference weight $\beta_i$. Participants slightly disagreed with this statement on average.}
    \label{fig:beta-preferences}
\end{figure}

For participants who went to the gym (37\%), we asked about their preferences over interventions to encourage gym attendance. For those who did not go to the gym, we asked about preferences over interventions to encourage their levels physical activity (in general). 
Participants were asked to \textit{``indicate the degree to which you think the following reminders or incentives would motivate you to increase your [gym attendance/levels of physical activity]''} on a 5-point Likert scale for five different intervention categories (financial incentives, messages that affirm your values, notifications to plan workouts, notifications to reflect on your number of gym visits per week). 
These categories were based on the interventions used in the gym attendance study~\citep{milkman2021megastudies}.
For each participant $i$, we converted these preferences into a vector $\boldsymbol{\alpha}_i \in \mathbb{R}^6$ by taking a softmax over ratings. 
Participants exhibited a great degree of diversity in their preferences (Figure \ref{fig:alpha-preferences}). On average, financial incentives were rated highest ratings and notifications were rated least lowest, though participants varied widely in their individual preferences. 

Participants were also asked about the extent to which they agree with the statement \textit{``If an artificial intelligence (AI) system were to make recommendations for increasing my [gym attendance/levels of physical activity], I would listen to the AI’s advice,''} which we converted to a scalar $\beta_i \in [0,1]$ indicating the degree to which an algorithm should prioritize participant preferences. As an example, if $\beta_i = 1$, this means participant $i$ would \textit{not} listen to an AI system, and thus an algorithm should prioritize the participant's reported preferences. 
Participants slightly disagreed with this statement on average (Figure \ref{fig:beta-preferences}).

\paragraph{Gym Attendance Dataset}
The gym attendance study was conducted in collaboration with 24 Hour Fitness, one of the largest gym chains in the United States. The study measures the impact of 54 different interventions on gym attendance over the course of a 4-week intervention period. The dataset contains information from over 60,000 participants across a wide variety of US states and age groups. 

For each participant, the dataset contains both demographic information (age, gender, US state of residence, new gym member status) as well as the weekly number of gym visits (an integer between 0-7) during the 4-week study period. Gym visits are also reported during a 10-week post-study period and some participants have up to a year of historical pre-study data. 
Participants were assigned to one of 54 treatments following a cohort-based randomization scheme. Participants were assigned to a cohort based on their study entry date and each cohort had different treatment assignment probabilities.

\paragraph{Gym Attendance Simulator}
\begin{figure*}
\begin{minipage}[t]{.48\textwidth}
\vspace{0pt}
\centering 
\includegraphics[width=\linewidth]{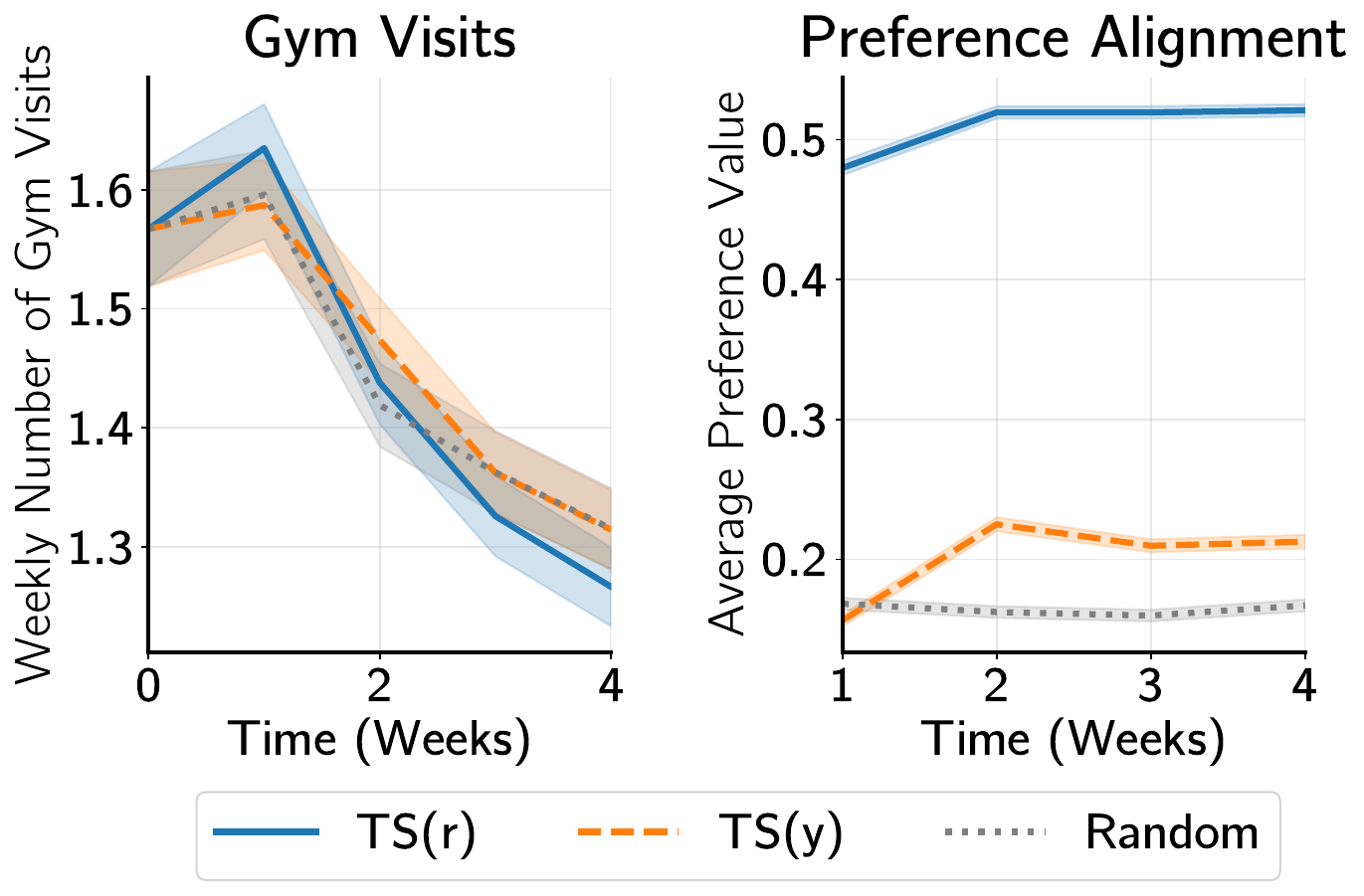}
  \caption{We compare the weekly number of gym visits (left) and average preference value $\boldsymbol{\alpha}(a)$ for the recommended actions (right) for several policies: our algorithm ($\textsf{TS}(r)$, {\color{blue} blue}), 
  Thompson sampling optimizing for gym visits ($\textsf{TS}(y)$, {\color{orange} orange}), and a random policy ({\color{gray} gray}). 
  We find that all policies achieve a similar number of gym visits, but only $\textsf{TS}(r)$ explicitly considers user preferences and achieves the highest preference alignment.
  Shaded area is standard error across 10 trials with $P=209$ participants each.
  }
  \label{fig:24h-reward-vs-utiliy}
\end{minipage}
\hfill
\begin{minipage}[t]{.48\textwidth}
\vspace{0pt}
\centering
\includegraphics[width=0.94\linewidth]{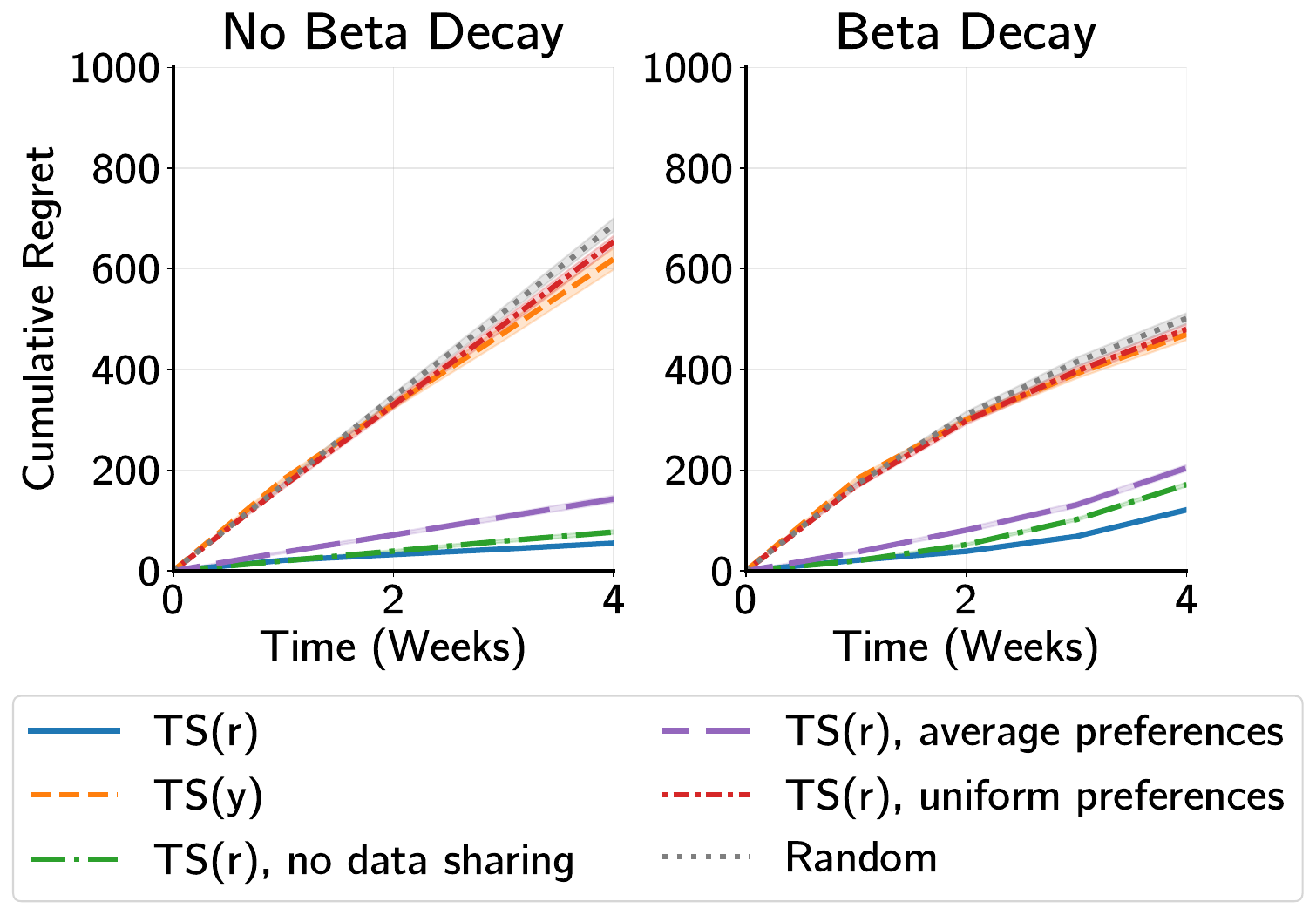}
  \caption{We compare our algorithm ($\textsf{TS}(r)$, {\color{blue} blue}) to TS optimizing for gym visits ($\textsf{TS}(y)$, {\color{orange} orange}), a random policy ({\color{gray} gray}), separate TS policies for each user ($\textsf{TS}(r)$, no data sharing, {\color{Green4} green}), TS using averaged values of $\boldsymbol{\alpha}$ and $\beta$ ($\textsf{TS}(r)$, average preferences, {\color{Purple1} purple}), and TS using uniform weights for $\boldsymbol{\alpha}$ and $\beta$ ($\textsf{TS}(r)$, uniform preferences, {\color{Red1} red}).
  We plot experiments without beta decay (left) and linear beta decay (right).
  Our algorithm achieves the lowest cumulative regret
  because it considers individualized preferences and shares data across users. 
  Shaded area is standard error across 10 trials with $P=209$ participants.}
  \label{fig:24h-cum-regret}
\end{minipage}
\vspace{-1em}
\end{figure*}
To use the gym attendance dataset as a bandit simulator, it is necessary to estimate counterfactual outcomes (i.e., how many times a participant would have gone to the gym if they had been assigned a different treatment).
We train a gradient-boosted random forest regression model to predict weekly gym visits $y_{i,t}$ given a featurization $\phi_{i,t}$ of the participant's history up until time $t$, their demographic information, and a one-hot encoding of $a_{i,t}$. When the treatment chosen by the bandit matches the treatment observed in the dataset, we use gym visits from the original dataset, not our regression model.

Since the dataset contains only one outcome variable (weekly gym visits) over which it is difficult to formulate a diversity of goals,\footnote{Under the assumption that most people would like to increase the number of gym visits, a policy optimizing for user-specific goals over gym visits would likely behave similarly to a policy optimizing directly for gym visits in the absence of other contextual information (e.g., notification count, visit duration, visit time, etc.).} we instead use preferences over the set of possible interventions to generate user-specific utility functions.
For simplicity, we categorized the original 54 interventions into the five (previously mentioned) intervention categories and restrict ourselves to participants who were assigned the intervention with the highest treatment effect within each category. 
We also include the placebo control condition. 
We found that this restricted action set facilitates bandit learning over the short, 4-week time horizon by reducing the dimensionality of our featurization and eliminating interventions with small treatment effects. Full details on our regression model can be found in Appendix \ref{appendix:24h:simulator}.

To match participant preferences from our online study to participants in the gym dataset, we use a randomized matching algorithm. 
We match each participant from the online study to a set of candidate participants in the test set of our gym attendance dataset based on gender, age (within 5 years), new member status, and average number of gym visits pre-intervention (within 0.5 visits). We treat participants in the online study who do not go to the gym as new members and set their pre-intervention attendance to 0. We randomly select amongst the candidate participants
that meet the matching criteria.

This matching procedure is not intended to suggest that the preferences of an online study participant are identical to the preferences of their matched counterpart in the gym attendance study based on the matching variables alone. In fact, preferences are likely to depend on a number of psychological factors that were not measured in the gym attendance dataset. 
Instead, we use randomized matching over many simulations to demonstrate that our algorithm can robustly optimize over any choice of preferences drawn from a realistic distribution.

\paragraph{Experiments}

Given our regression-based step count simulator and matched participant preferences, we construct the following reward function,
\begin{equation}
\hspace{-0.4em}
    r_{i,t}(s_{i,t}, a_{i,t}) = \beta_{i,t}\gamma  \boldsymbol{\alpha}_i(a_{i,t}) + (1 - \beta_{i,t}) y(s_{i,t}, a_{i,t}).
    \label{eq:reward-24h}
\end{equation}
This reward function can be interpreted as weighted sum of two utility functions (Eq.~\ref{eq:reward}): (1) $U_{i,t,1}(y_{i,t}, \boldsymbol{\phi}_{i,t}) = \gamma \boldsymbol{\alpha}_i(a_{i,t})$ rewards actions that maximize participant preferences and (2) $U_{i,t,2}(y_{i,t}, \boldsymbol{\phi}_{i,t}) = y_{i,t}$ rewards actions that maximize gym visits. 
$\gamma = 7.59$ is a scaling constant that ensures that the expectation of both utility functions is equal, set to the mean of all $y$ in the test set divided by the mean of all $\boldsymbol{\alpha}_i$ in our survey. 
The degree to which an algorithm should prioritize preferences over gym visits is determined by $w_{i,t,1} = \beta_{i,t}$ and $w_{i,t,2} = 1 - \beta_{i,t}$.
The first utility function is an action-dependent constant and the second is the identity, both of which are Lipschitz in $y$.

We compare gym visits and preference alignment (the average value of $\boldsymbol{\alpha}(a)$ for all actions pulled by the policy) for TS optimizing for reward $r_{i, t}$ ($\textsf{TS}(r)$, Algorithm \ref{alg:ts_aug}), TS optimizing for gym visits $y_{i, t}$ ($\textsf{TS}(y)$), and a random policy (Figure \ref{fig:24h-reward-vs-utiliy}).
All policies yield a comparable number of gym visits. The downward trend is a general trend observed in this dataset (and behavior change studies more broadly), where treatment effect is highest initially and tapers over time.
Meanwhile, Thompson sampling optimizing for reward selects actions that are substantially more aligned with participant preferences. 

Next, we simulate a 4-week study with $P=209$ participants and plot cumulative regret for several algorithms: 
$\textsf{TS}(r)$, $\textsf{TS}(y)$, an independent $\textsf{TS}(r)$ policy for each user that does not share data, a $\textsf{TS}(r)$ policy that uses averaged values of $\boldsymbol{\alpha}$ and $\beta$ instead of personalized preferences, a $\textsf{TS}(r)$ policy that uses uniform values for $\boldsymbol{\alpha}$ and $\beta$, and a random policy.
In one version of the simulation, we linearly decay $\beta$ to zero over the course of the study, 
setting $\beta_{i,1} = \beta_i$ and $\beta_{i,T} = 0$.
Beta decay incentivizes the policy to respect participant preferences early in the study, while placing a higher weight on interventions that encourage gym visits later in the study. In another version, $\beta$ is held constant across all timesteps. We assign treatment in parallel, as in the previous simulator.
In both versions, we find that TS optimizing for $r_{i, t}$ achieves the lowest cumulative regret, where the regret boost is due to both data sharing and recommending actions that align with participant preferences (Figure \ref{fig:24h-cum-regret}). TS optimizing for gym visits performs comparably to random assignment due to its inability to consider participant preferences.

In light of prior evidence that personalization affects mHealth app adherence~\citep{yang2020factors} and well-documented literature on algorithm aversion~\citep{dietvorst2015algorithm}, we believe that respecting participant preferences, particularly early in a study, can lead to increased trust and adherence.
The original gym attendance study did not assign interventions with regard for participant preferences and we suspect that randomly assigned interventions may have impacted engagement and adherence (e.g., the downward trend in Figure \ref{fig:24h-reward-vs-utiliy}). In practice, interventions that are preference-aligned may inherently be more effective \textit{because} they are preference-aligned. We are unable to capture these compound effects in our simulator because participant preferences are divorced from their dynamics.
\section{Discussion}
\label{sec:discussion}
In this work, we introduce a modification to Thompson sampling that allows us to effectively optimize for user-specific goals, achieving similar cumulative regret bounds to existing approaches. In synthetic and semi-synthetic experiments, our algorithm achieves a cumulative regret speedup in comparison to policies that are trained separately for each user or do not account for user-specific goals.

Our long-term goal is to evaluate our algorithm's performance in a field study measuring physical activity behavior change.
Before we evaluate our method in a field study, there are are a number of practical challenges that must be addressed. First, our method assumes that goals and preferences can be quantified by end-users. However, goals are often easier to formulate at a higher level (e.g., \textit{``I want to feel healthier in my body''}) and people will likely require assistance in translating these abstract goals into the measurable, quantitative formulation our method requires~\citep{goals_goalsetting}. 
Goals 
are also dynamic and frequently shift over time~\citep{goals_dynamic}, which our algorithm supports in theory but was not evaluated in our experiments. 
Similarly, our preference elicitation method may not capture participants' true underlying preferences, which can also change over time~\citep{preferences_hard, preferences_stress}, and further work is required to establish robust preference elicitation methods. 
Lastly, the space of possible health behavior change is enormous, and our algorithm's performance scales linearly with the dimensionality of the featurization. Thus, it may be necessary to restrict the action space for each user.
We are hopeful that large language models (LLMs) may assist in many of these challenges. 
However, we believe that statistical algorithms like ours will be more effective at estimating treatment effect and quantifying uncertainty to guide adaptive experimentation in the limited-data settings common in mobile health studies. An LLM could feasibly be used to ``warm-start'' our algorithm, and we are particularly interested in exploring the intersection of LLMs and statistical learning algorithms like ours.

While our linear Thompson sampling algorithm performed well in our experiments, 
different deployment scenarios and outcome variables may require more complex modeling choices. 
For instance, it is possible to relax our linearity assumptions, e.g., using Gaussian processes~\citep{chowdhury2017kernelized} or semi-parametric models~\citep{greenewald2017action, krishnamurthy2018semiparametric}, and explicitly account for treatment effect heterogeneity, e.g., using hierarchical~\citep{hong2022hierarchical} or mixed-effects models~\citep{tomkins2020intelligentpooling}. 
Moreover, while our bandit model has sample complexity benefits, it does not explicitly account for non-stationarity and habituation effects. It may be more appropriate to model long-term treatment effect dynamics using an MDP model, which could also be learned using Thompson sampling~\citep{osband2013more}.

Towards expanding our theoretical results, our algorithm and regret bounds assume that each action is assigned sequentially (i.e., the model is updated between assigning treatment to user $i$ and $i+1$), though in practice, treatment is likely assigned in parallel. 
Future work could investigate regret bounds for parallel assignment. 
Further, it may be possible to achieve similar regret guarantees for non-Lipschitz utility functions, e.g., indicator functions with a finite number of discontinuities. It may also prove useful to derive simple regret bounds to investigate whether the presence of user-specific goals can change exploration strategies in a pure exploration setting.

Lastly, we mention that our algorithm may prove useful in several other domains beyond health behavior change. For example, in education, students have personalized learning goals for which an algorithm could recommend various learning activities. For financial savings, users may have goals related to different spending categories and algorithms could propose various saving interventions. We welcome further discussion on different application domains that could benefit from our method. 

\acks{The authors would like to thank Scott Fleming, Jiayu Yao, Yash Chandak, and Jonathan Lee for feedback. AM was supported in part by a Stanford Engineering fellowship. AM and BEE were funded in part by Helmsley Trust grant AWD1006624, NIH NCI 5U2CCA233195, NSF CAREER AWD1005627, and NIH NHGRI R01 HG012967. BEE is a CIFAR Fellow in the Multiscale Human Program.}
\bibliography{references}

\begin{thebibliography}{61}
\providecommand{\natexlab}[1]{#1}
\providecommand{\url}[1]{\texttt{#1}}
\expandafter\ifx\csname urlstyle\endcsname\relax
  \providecommand{\doi}[1]{doi: #1}\else
  \providecommand{\doi}{doi: \begingroup \urlstyle{rm}\Url}\fi

\bibitem[Agrawal and Goyal(2013)]{agrawal2013thompson}
Shipra Agrawal and Navin Goyal.
\newblock Thompson sampling for contextual bandits with linear payoffs.
\newblock In \emph{International conference on machine learning}, pages 127--135. PMLR, 2013.

\bibitem[Baek et~al.(2023)Baek, Boutilier, Farias, Jonasson, and Yoeli]{baek2023policy}
Jackie Baek, Justin~J. Boutilier, Vivek~F. Farias, Jonas~Oddur Jonasson, and Erez Yoeli.
\newblock Policy optimization for personalized interventions in behavioral health, 2023.

\bibitem[Bertsimas et~al.(2022)Bertsimas, Klasnja, Murphy, and Na]{data_driven_interpretable}
Dimitris Bertsimas, Predrag Klasnja, Susan Murphy, and Liangyuan Na.
\newblock Data-driven interpretable policy construction for personalized mobile health.
\newblock In \emph{2022 IEEE International Conference on Digital Health (ICDH)}, pages 13--22, 2022.
\newblock \doi{10.1109/ICDH55609.2022.00010}.

\bibitem[Bickmore et~al.(2011)Bickmore, Schulman, and Sidner]{bickmore2011reusable}
Timothy~W Bickmore, Daniel Schulman, and Candace~L Sidner.
\newblock A reusable framework for health counseling dialogue systems based on a behavioral medicine ontology.
\newblock \emph{Journal of biomedical informatics}, 44\penalty0 (2):\penalty0 183--197, 2011.

\bibitem[Buttorff et~al.(2017)Buttorff, Ruder, Bauman, et~al.]{buttorff2017multiple}
Christine Buttorff, Teague Ruder, Melissa Bauman, et~al.
\newblock \emph{Multiple chronic conditions in the United States}, volume~10.
\newblock Rand Santa Monica, CA, 2017.

\bibitem[Chowdhury and Gopalan(2017)]{chowdhury2017kernelized}
Sayak~Ray Chowdhury and Aditya Gopalan.
\newblock On kernelized multi-armed bandits.
\newblock In \emph{International Conference on Machine Learning}, pages 844--853. PMLR, 2017.

\bibitem[Deshmukh et~al.(2017)Deshmukh, Dogan, and Scott]{deshmukh2017multitask}
Aniket~Anand Deshmukh, Urun Dogan, and Clayton Scott.
\newblock Multi-task learning for contextual bandits, 2017.

\bibitem[Dietvorst et~al.(2015)Dietvorst, Simmons, and Massey]{dietvorst2015algorithm}
Berkeley~J Dietvorst, Joseph~P Simmons, and Cade Massey.
\newblock Algorithm aversion: people erroneously avoid algorithms after seeing them err.
\newblock \emph{Journal of Experimental Psychology: General}, 144\penalty0 (1):\penalty0 114, 2015.

\bibitem[Dimakopoulou and Roy(2018)]{dimakopoulou2018coordinated}
Maria Dimakopoulou and Benjamin~Van Roy.
\newblock Coordinated exploration in concurrent reinforcement learning, 2018.

\bibitem[Domin et~al.(2021)Domin, Spruijt-Metz, Theisen, Ouzzahra, V{\"o}gele, et~al.]{domin2021smartphone}
Alex Domin, Donna Spruijt-Metz, Daniel Theisen, Yacine Ouzzahra, Claus V{\"o}gele, et~al.
\newblock Smartphone-based interventions for physical activity promotion: scoping review of the evidence over the last 10 years.
\newblock \emph{JMIR mHealth and uHealth}, 9\penalty0 (7):\penalty0 e24308, 2021.

\bibitem[Ekhtiar et~al.(2023)Ekhtiar, Karahano\u{g}lu, Gouveia, and Ludden]{goals_goalsetting}
Tina Ekhtiar, Arma\u{g}an Karahano\u{g}lu, R\'{u}ben Gouveia, and Geke Ludden.
\newblock Goals for goal setting: A scoping review on personal informatics.
\newblock In \emph{Proceedings of the 2023 ACM Designing Interactive Systems Conference}, DIS '23, page 2625–2641, New York, NY, USA, 2023. Association for Computing Machinery.
\newblock ISBN 9781450398930.
\newblock \doi{10.1145/3563657.3596087}.
\newblock URL \url{https://doi.org/10.1145/3563657.3596087}.

\bibitem[Epton et~al.(2017)Epton, Currie, and Armitage]{epton2017unique}
Tracy Epton, Sinead Currie, and Christopher~J Armitage.
\newblock Unique effects of setting goals on behavior change: Systematic review and meta-analysis.
\newblock \emph{Journal of consulting and clinical psychology}, 85\penalty0 (12):\penalty0 1182, 2017.

\bibitem[Garivier and Moulines(2008)]{garivier2008upperconfidence}
Aurélien Garivier and Eric Moulines.
\newblock On upper-confidence bound policies for non-stationary bandit problems, 2008.

\bibitem[Gentile et~al.(2014)Gentile, Li, and Zappella]{gentile2014online}
Claudio Gentile, Shuai Li, and Giovanni Zappella.
\newblock Online clustering of bandits, 2014.

\bibitem[Gentile et~al.(2017)Gentile, Li, Kar, Karatzoglou, Etrue, and Zappella]{gentile2017contextdependent}
Claudio Gentile, Shuai Li, Purushottam Kar, Alexandros Karatzoglou, Evans Etrue, and Giovanni Zappella.
\newblock On context-dependent clustering of bandits, 2017.

\bibitem[Greenewald et~al.(2017)Greenewald, Tewari, Murphy, and Klasnja]{greenewald2017action}
Kristjan Greenewald, Ambuj Tewari, Susan Murphy, and Predag Klasnja.
\newblock Action centered contextual bandits.
\newblock \emph{Advances in neural information processing systems}, 30, 2017.

\bibitem[Guo and Brunskill(2015)]{Guo_Brunskill_2015}
Zhaohan Guo and Emma Brunskill.
\newblock Concurrent pac rl.
\newblock \emph{Proceedings of the AAAI Conference on Artificial Intelligence}, 29\penalty0 (1), Feb. 2015.
\newblock \doi{10.1609/aaai.v29i1.9585}.
\newblock URL \url{https://ojs.aaai.org/index.php/AAAI/article/view/9585}.

\bibitem[Hagger(2019)]{hagger2019habit}
Martin~S Hagger.
\newblock Habit and physical activity: Theoretical advances, practical implications, and agenda for future research.
\newblock \emph{Psychology of Sport and Exercise}, 42:\penalty0 118--129, 2019.

\bibitem[Hall et~al.(2010)Hall, Crowley, Bosworth, Howard, and Morey]{goals_dynamic}
Katherine Hall, Gail Crowley, Hayden Bosworth, Teresa Howard, and Miriam Morey.
\newblock Individual progress toward self-selected goals among older adults enrolled in a physical activity counseling intervention.
\newblock \emph{Journal of aging and physical activity}, 18:\penalty0 439--50, 10 2010.
\newblock \doi{10.1123/japa.18.4.439}.

\bibitem[Hariri et~al.(2015)Hariri, Mobasher, and Burke]{Hariri2015AdaptingTU}
Negar Hariri, Bamshad Mobasher, and R.~Burke.
\newblock Adapting to user preference changes in interactive recommendation.
\newblock In \emph{International Joint Conference on Artificial Intelligence}, 2015.

\bibitem[Hicks et~al.(2023)Hicks, Boswell, Althoff, Crum, Ku, Landay, Moya, Murnane, Snyder, King, et~al.]{hicks2023leveraging}
Jennifer~L Hicks, Melissa~A Boswell, Tim Althoff, Alia~J Crum, Joy~P Ku, James~A Landay, Paula~ML Moya, Elizabeth~L Murnane, Michael~P Snyder, Abby~C King, et~al.
\newblock Leveraging mobile technology for public health promotion: A multidisciplinary perspective.
\newblock \emph{Annual Review of Public Health}, 44:\penalty0 131--150, 2023.

\bibitem[Hong et~al.(2022)Hong, Kveton, Zaheer, and Ghavamzadeh]{hong2022hierarchical}
Joey Hong, Branislav Kveton, Manzil Zaheer, and Mohammad Ghavamzadeh.
\newblock Hierarchical bayesian bandits, 2022.

\bibitem[Klasnja et~al.(2015)Klasnja, Hekler, Shiffman, Boruvka, Almirall, Tewari, and Murphy]{klasnja2015microrandomized}
Predrag Klasnja, Eric~B Hekler, Saul Shiffman, Audrey Boruvka, Daniel Almirall, Ambuj Tewari, and Susan~A Murphy.
\newblock Microrandomized trials: An experimental design for developing just-in-time adaptive interventions.
\newblock \emph{Health Psychology}, 34\penalty0 (S):\penalty0 1220, 2015.

\bibitem[Klasnja et~al.(2019)Klasnja, Smith, Seewald, Lee, Hall, Luers, Hekler, and Murphy]{heartsteps}
Predrag Klasnja, Shawna Smith, Nicholas~J Seewald, Andy Lee, Kelly Hall, Brook Luers, Eric~B Hekler, and Susan~A Murphy.
\newblock Efficacy of contextually tailored suggestions for physical activity: a micro-randomized optimization trial of heartsteps.
\newblock \emph{Annals of Behavioral Medicine}, 53\penalty0 (6):\penalty0 573--582, 2019.

\bibitem[Krishnamurthy et~al.(2018)Krishnamurthy, Wu, and Syrgkanis]{krishnamurthy2018semiparametric}
Akshay Krishnamurthy, Zhiwei~Steven Wu, and Vasilis Syrgkanis.
\newblock Semiparametric contextual bandits.
\newblock In \emph{International Conference on Machine Learning}, pages 2776--2785. PMLR, 2018.

\bibitem[Kveton et~al.(2021)Kveton, Konobeev, Zaheer, wei Hsu, Mladenov, Boutilier, and Szepesvari]{kveton2021metathompson}
Branislav Kveton, Mikhail Konobeev, Manzil Zaheer, Chih wei Hsu, Martin Mladenov, Craig Boutilier, and Csaba Szepesvari.
\newblock Meta-thompson sampling, 2021.

\bibitem[Lattimore and Szepesv{\'a}ri(2020)]{lattimore2020bandit}
Tor Lattimore and Csaba Szepesv{\'a}ri.
\newblock \emph{Bandit algorithms}.
\newblock Cambridge University Press, 2020.

\bibitem[Liao et~al.(2015)Liao, Klasnja, Tewari, and Murphy]{effect_changes}
Peng Liao, Predrag Klasnja, Ambuj Tewari, and Susan Murphy.
\newblock Sample size calculations for micro-randomized trials in mhealth.
\newblock \emph{Statistics in medicine}, 35, 12 2015.
\newblock \doi{10.1002/sim.6847}.

\bibitem[Liao et~al.(2016)Liao, Klasnja, Tewari, and Murphy]{liao2016sample}
Peng Liao, Predrag Klasnja, Ambuj Tewari, and Susan~A Murphy.
\newblock Sample size calculations for micro-randomized trials in mhealth.
\newblock \emph{Statistics in medicine}, 35\penalty0 (12):\penalty0 1944--1971, 2016.

\bibitem[Liao et~al.(2019)Liao, Greenewald, Klasnja, and Murphy]{liao2019personalized}
Peng Liao, Kristjan Greenewald, Predrag Klasnja, and Susan Murphy.
\newblock Personalized heartsteps: A reinforcement learning algorithm for optimizing physical activity, 2019.

\bibitem[Locke and Latham(2002)]{locke2002building}
Edwin~A Locke and Gary~P Latham.
\newblock Building a practically useful theory of goal setting and task motivation: A 35-year odyssey.
\newblock \emph{American psychologist}, 57\penalty0 (9):\penalty0 705, 2002.

\bibitem[Luo et~al.(2019)Luo, Wei, Agarwal, and Langford]{luo2019efficient}
Haipeng Luo, Chen-Yu Wei, Alekh Agarwal, and John Langford.
\newblock Efficient contextual bandits in non-stationary worlds, 2019.

\bibitem[Mair et~al.(2012)Mair, May, O'Donnell, Finch, Sullivan, and Murray]{preferences_hard}
Frances Mair, Carl May, Kate O'Donnell, Tracy Finch, Frank Sullivan, and Elizabeth Murray.
\newblock Factors that promote or inhibit the implementation of e-health systems: An explanatory systematic review.
\newblock \emph{Bulletin of the World Health Organization}, 90:\penalty0 357--64, 05 2012.
\newblock \doi{10.2471/BLT.11.099424}.

\bibitem[McEwan et~al.(2016)McEwan, Harden, Zumbo, Sylvester, Kaulius, Ruissen, Dowd, and Beauchamp]{mcewan2016effectiveness}
Desmond McEwan, Samantha~M Harden, Bruno~D Zumbo, Benjamin~D Sylvester, Megan Kaulius, Geralyn~R Ruissen, A~Justine Dowd, and Mark~R Beauchamp.
\newblock The effectiveness of multi-component goal setting interventions for changing physical activity behaviour: a systematic review and meta-analysis.
\newblock \emph{Health psychology review}, 10\penalty0 (1):\penalty0 67--88, 2016.

\bibitem[Milkman et~al.(2021)Milkman, Gromet, Ho, Kay, Lee, Pandiloski, Park, Rai, Bazerman, Beshears, Bonacorsi, Camerer, Chang, Chapman, Cialdini, Dai, Eskreis-Winkler, Fishbach, Gross, and Duckworth]{milkman2021megastudies}
Katherine Milkman, Dena Gromet, Hung Ho, Joseph Kay, Timothy Lee, Pepi Pandiloski, Yeji Park, Aneesh Rai, Max Bazerman, John Beshears, Lauri Bonacorsi, Colin Camerer, Edward Chang, Gretchen Chapman, Robert Cialdini, Hengchen Dai, Lauren Eskreis-Winkler, Ayelet Fishbach, James Gross, and Angela Duckworth.
\newblock Megastudies improve the impact of applied behavioural science.
\newblock \emph{Nature}, 600, 12 2021.
\newblock \doi{10.1038/s41586-021-04128-4}.

\bibitem[Mintz et~al.(2019)Mintz, Aswani, Kaminsky, Flowers, and Fukuoka]{mintz2019nonstationary}
Yonatan Mintz, Anil Aswani, Philip Kaminsky, Elena Flowers, and Yoshimi Fukuoka.
\newblock Non-stationary bandits with habituation and recovery dynamics, 2019.

\bibitem[Mitchell et~al.(2021)Mitchell, Maimone, Cassells, Tobin, Davidson, Smaldone, and Mamykina]{mitchell2021automated}
Elliot~G Mitchell, Rosa Maimone, Andrea Cassells, Jonathan~N Tobin, Patricia Davidson, Arlene~M Smaldone, and Lena Mamykina.
\newblock Automated vs. human health coaching: exploring participant and practitioner experiences.
\newblock \emph{Proceedings of the ACM on human-computer interaction}, 5\penalty0 (CSCW1):\penalty0 1--37, 2021.

\bibitem[Nahum-Shani et~al.(2016)Nahum-Shani, Smith, Spring, Collins, Witkiewitz, Tewari, and Murphy]{jitai}
Inbal Nahum-Shani, Shawna Smith, Bonnie Spring, Linda Collins, Katie Witkiewitz, Ambuj Tewari, and Susan Murphy.
\newblock Just-in-time adaptive interventions (jitais) in mobile health: Key components and design principles for ongoing health behavior support.
\newblock \emph{Annals of Behavioral Medicine}, 52, 09 2016.
\newblock \doi{10.1007/s12160-016-9830-8}.

\bibitem[Olsen(2014)]{olsen2014health}
Jeanette~M Olsen.
\newblock Health coaching: a concept analysis.
\newblock In \emph{Nursing forum}, volume~49, pages 18--29. Wiley Online Library, 2014.

\bibitem[Olsen and Nesbitt(2010)]{olsen2010health}
Jeanette~M Olsen and Bonnie~J Nesbitt.
\newblock Health coaching to improve healthy lifestyle behaviors: an integrative review.
\newblock \emph{American journal of health promotion}, 25\penalty0 (1):\penalty0 e1--e12, 2010.

\bibitem[Osband et~al.(2013)Osband, Russo, and Van~Roy]{osband2013more}
Ian Osband, Daniel Russo, and Benjamin Van~Roy.
\newblock (more) efficient reinforcement learning via posterior sampling.
\newblock \emph{Advances in Neural Information Processing Systems}, 26, 2013.

\bibitem[Paredes et~al.(2014)Paredes, Gilad-Bachrach, Czerwinski, Roseway, Rowan, and Hernandez]{poptherapy}
Pablo Paredes, Ran Gilad-Bachrach, Mary Czerwinski, Asta Roseway, Kael Rowan, and Javier Hernandez.
\newblock Poptherapy: coping with stress through pop-culture.
\newblock In \emph{Proceedings of the 8th International Conference on Pervasive Computing Technologies for Healthcare}, PervasiveHealth '14, page 109–117, Brussels, BEL, 2014. ICST (Institute for Computer Sciences, Social-Informatics and Telecommunications Engineering).
\newblock ISBN 9781631900112.
\newblock \doi{10.4108/icst.pervasivehealth.2014.255070}.
\newblock URL \url{https://doi.org/10.4108/icst.pervasivehealth.2014.255070}.

\bibitem[Pazis and Parr(2016)]{Pazis_Parr_2016}
Jason Pazis and Ronald Parr.
\newblock Efficient pac-optimal exploration in concurrent, continuous state mdps with delayed updates.
\newblock \emph{Proceedings of the AAAI Conference on Artificial Intelligence}, 30\penalty0 (1), Mar. 2016.
\newblock \doi{10.1609/aaai.v30i1.10307}.
\newblock URL \url{https://ojs.aaai.org/index.php/AAAI/article/view/10307}.

\bibitem[Peng et~al.(2021)Peng, Li, Kononova, Cotten, Kamp, Bowen, et~al.]{peng2021habit}
Wei Peng, Lin Li, Anastasia Kononova, Shelia Cotten, Kendra Kamp, Marie Bowen, et~al.
\newblock Habit formation in wearable activity tracker use among older adults: qualitative study.
\newblock \emph{JMIR mHealth and uHealth}, 9\penalty0 (1):\penalty0 e22488, 2021.

\bibitem[Rabbi et~al.(2015)Rabbi, Aung, Zhang, and Choudhury]{Rabbi2015MyBehaviorAP}
Mashfiqui Rabbi, M.~S.~Hane Aung, Mi~Zhang, and Tanzeem Choudhury.
\newblock Mybehavior: automatic personalized health feedback from user behaviors and preferences using smartphones.
\newblock \emph{Proceedings of the 2015 ACM International Joint Conference on Pervasive and Ubiquitous Computing}, 2015.
\newblock URL \url{https://api.semanticscholar.org/CorpusID:207225576}.

\bibitem[Roijers et~al.(2021)Roijers, Zintgraf, Libin, Reymond, Bargiacchi, and Nowe]{interactive_multiobjective}
Diederik Roijers, Luisa Zintgraf, Pieter Libin, Mathieu Reymond, Eugenio Bargiacchi, and Ann Nowe.
\newblock \emph{Interactive Multi-objective Reinforcement Learning in Multi-armed Bandits with Gaussian Process Utility Models}, pages 463--478.
\newblock 02 2021.
\newblock ISBN 978-3-030-67663-6.
\newblock \doi{10.1007/978-3-030-67664-3_28}.

\bibitem[Ruggeri et~al.(2023)Ruggeri, Benzerga, Verra, and Folke]{ruggeri_benzerga_verra_folke_2023}
Kai Ruggeri, Amel Benzerga, Sanne Verra, and Tomas Folke.
\newblock A behavioral approach to personalizing public health.
\newblock \emph{Behavioural Public Policy}, 7\penalty0 (2):\penalty0 457–469, 2023.
\newblock \doi{10.1017/bpp.2020.31}.

\bibitem[Rutjes et~al.(2019)Rutjes, Willemsen, and IJsselsteijn]{rutjes2019beyond}
Heleen Rutjes, Martijn~C Willemsen, and Wijnand~A IJsselsteijn.
\newblock Beyond behavior: the coach's perspective on technology in health coaching.
\newblock In \emph{Proceedings of the 2019 CHI Conference on Human Factors in Computing Systems}, pages 1--14, 2019.

\bibitem[Shcherbina et~al.(2019)Shcherbina, Hershman, Lazzeroni, King, O'Sullivan, Hekler, Moayedi, Pavlovic, Waggott, Sharma, Yeung, Christle, Wheeler, McConnell, Harrington, and Ashley]{SHCHERBINA2019e344}
Anna Shcherbina, Steven~G Hershman, Laura Lazzeroni, Abby~C King, Jack~W O'Sullivan, Eric Hekler, Yasbanoo Moayedi, Aleksandra Pavlovic, Daryl Waggott, Abhinav Sharma, Alan Yeung, Jeffrey~W Christle, Matthew~T Wheeler, Michael~V McConnell, Robert~A Harrington, and Euan~A Ashley.
\newblock The effect of digital physical activity interventions on daily step count: a randomised controlled crossover substudy of the myheart counts cardiovascular health study.
\newblock \emph{The Lancet Digital Health}, 1\penalty0 (7):\penalty0 e344--e352, 2019.
\newblock ISSN 2589-7500.
\newblock \doi{https://doi.org/10.1016/S2589-7500(19)30129-3}.

\bibitem[Silver et~al.(2013)Silver, Newnham, Barker, Weller, and McFall]{pmlr-v28-silver13}
David Silver, Leonard Newnham, David Barker, Suzanne Weller, and Jason McFall.
\newblock Concurrent reinforcement learning from customer interactions.
\newblock In Sanjoy Dasgupta and David McAllester, editors, \emph{Proceedings of the 30th International Conference on Machine Learning}, volume~28 of \emph{Proceedings of Machine Learning Research}, pages 924--932, Atlanta, Georgia, USA, 17--19 Jun 2013. PMLR.
\newblock URL \url{https://proceedings.mlr.press/v28/silver13.html}.

\bibitem[Tomkins et~al.(2020)Tomkins, Liao, Klasnja, and Murphy]{tomkins2020intelligentpooling}
Sabina Tomkins, Peng Liao, Predrag Klasnja, and Susan Murphy.
\newblock Intelligentpooling: Practical thompson sampling for mhealth, 2020.

\bibitem[Torkamaan and Ziegler(2019)]{preferences_stress}
Helma Torkamaan and J\"{u}rgen Ziegler.
\newblock Rating-based preference elicitation for recommendation of stress intervention.
\newblock In \emph{Proceedings of the 27th ACM Conference on User Modeling, Adaptation and Personalization}, UMAP '19, page 46–50, New York, NY, USA, 2019. Association for Computing Machinery.
\newblock ISBN 9781450360210.
\newblock \doi{10.1145/3320435.3324990}.
\newblock URL \url{https://doi.org/10.1145/3320435.3324990}.

\bibitem[Wang et~al.(2021)Wang, Zhang, Kröse, and Hoof]{notification_burden}
Shihan Wang, Chao Zhang, Ben Kröse, and Herke Hoof.
\newblock Optimizing adaptive notifications in mobile health interventions systems: Reinforcement learning from a data-driven behavioral simulator.
\newblock \emph{Journal of Medical Systems}, 45, 12 2021.
\newblock \doi{10.1007/s10916-021-01773-0}.

\bibitem[Watson et~al.(2022)Watson, Carlson, Loustalot, Town, Eke, Thomas, and Greenlund]{watson2022chronic}
Kathleen~B Watson, Susan~A Carlson, Fleetwood Loustalot, Machell Town, Paul~I Eke, Craig~W Thomas, and Kurt~J Greenlund.
\newblock Chronic conditions among adults aged 18–34 years—united states, 2019.
\newblock \emph{Morbidity and mortality weekly report}, 71\penalty0 (30):\penalty0 964, 2022.

\bibitem[Wolever et~al.(2013)Wolever, Simmons, Sforzo, Dill, Kaye, Bechard, Southard, Kennedy, Vosloo, and Yang]{wolever2013systematic}
Ruth~Q Wolever, Leigh~Ann Simmons, Gary~A Sforzo, Diana Dill, Miranda Kaye, Elizabeth~M Bechard, Mary~Elaine Southard, Mary Kennedy, Justine Vosloo, and Nancy Yang.
\newblock A systematic review of the literature on health and wellness coaching: defining a key behavioral intervention in healthcare.
\newblock \emph{Global advances in health and medicine}, 2\penalty0 (4):\penalty0 38--57, 2013.

\bibitem[{World Health Organization}(2022)]{who_pa}
{World Health Organization}.
\newblock Physical activity, 2022.
\newblock URL \url{https://www.who.int/en/news-room/fact-sheets/detail/physical-activity}.

\bibitem[Wu et~al.(2018)Wu, Iyer, and Wang]{Wu_2018}
Qingyun Wu, Naveen Iyer, and Hongning Wang.
\newblock Learning contextual bandits in a non-stationary environment.
\newblock In \emph{The 41st International ACM SIGIR Conference on Research: Development in Information Retrieval}, SIGIR ’18. ACM, June 2018.
\newblock \doi{10.1145/3209978.3210051}.
\newblock URL \url{http://dx.doi.org/10.1145/3209978.3210051}.

\bibitem[Yang and Van~Stee(2019)]{yang2019comparative}
Qinghua Yang and Stephanie~K Van~Stee.
\newblock The comparative effectiveness of mobile phone interventions in improving health outcomes: meta-analytic review.
\newblock \emph{JMIR mHealth and uHealth}, 7\penalty0 (4):\penalty0 e11244, 2019.

\bibitem[Yang et~al.(2020)Yang, Ma, Zhao, and Kankanhalli]{yang2020factors}
Xiaotian Yang, Lin Ma, Xi~Zhao, and Atreyi Kankanhalli.
\newblock Factors influencing user’s adherence to physical activity applications: A scoping literature review and future directions.
\newblock \emph{International Journal of Medical Informatics}, 134:\penalty0 104039, 2020.

\bibitem[Yao et~al.(2021)Yao, Brunskill, Pan, Murphy, and Doshi-Velez]{yao2021power}
Jiayu Yao, Emma Brunskill, Weiwei Pan, Susan Murphy, and Finale Doshi-Velez.
\newblock Power constrained bandits.
\newblock In \emph{Proceedings of the 6th Machine Learning for Healthcare Conference}, pages 209--259, 2021.

\bibitem[Yom-Tov et~al.(2017)Yom-Tov, Feraru, Kozdoba, Mannor, Tennenholtz, and Hochberg]{diabetes_activity}
Elad Yom-Tov, Guy Feraru, Mark Kozdoba, Shie Mannor, Moshe Tennenholtz, and Irit Hochberg.
\newblock Encouraging physical activity in patients with diabetes: Intervention using a reinforcement learning system.
\newblock \emph{Journal of Medical Internet Research}, 19:\penalty0 e338, 10 2017.
\newblock \doi{10.2196/jmir.7994}.

\end{thebibliography}
\onecolumn 
\appendix

\section{Bayesian Regret Proof}
\label{appendix:regret}
\newcommand*{\vertbar}{\rule[-1ex]{0.5pt}{2.5ex}}
\newcommand*{\horzbar}{\rule[.5ex]{2.5ex}{0.5pt}}
\setlength\parindent{0pt}
The proof structure largely follows \cite{lattimore2020bandit} Theorem 36.4, with modifications to account for our Lipschitz utility transformations. \\

\begin{customthm}{1}
After running Algorithm \ref{alg:ts_aug} for $N=P \cdot T$ samples, where $P$ is the number of participants in the cohort and $T$ is the time horizon, we achieves Bayesian cumulative regret on the order of
\begin{equation*}
    \text{BR}(N) \leq O(Ld \sqrt{N \log(NM)\log(N/d)})
\end{equation*}
\end{customthm}

\textit{Proof:}
For simplicity, we define $n = P t + i$ to convert the nested index $(i,t)$ to a single index. 
Using this notation, recall that
\begin{align}
    r_n(s, a) &= \sum_{k=1}^K w_{n,k} \cdot U_{n,k}\Big([\mathbf{y}(s, a)]_{m_k}, \boldsymbol{\phi}(s, a)\Big), \quad
    \text{where } [\mathbf{y}(s, a)]_m = \boldsymbol{\theta}_m^\top \boldsymbol{\phi}(s,a) + \varepsilon, \quad
    \sum_{k=1}^K w_{n,k} = 1
\end{align}

Writing $y_{n,m} \triangleq [\mathbf{y}(s_n, a_n)]_m$ and $\boldsymbol{\phi}_n \triangleq \boldsymbol{\phi}(s_n, a_n)$, define
\begin{align}
    \hat{\boldsymbol{\theta}}_{n,m} &= \Sigma_n^{-1} \Phi_n^\top Y_{n, m}, \quad
    \Phi_n = \begin{bmatrix}
        \horzbar & \boldsymbol{\phi}_1^\top & \horzbar \\
        & \vdots & \\
        \horzbar & \boldsymbol{\phi}_n^\top & \horzbar \\
    \end{bmatrix}, \quad
    Y_{n,m} = \begin{bmatrix}
        y_{1,m} \\
        \vdots \\
        y_{n,m}
    \end{bmatrix}, \quad
    \Sigma_n = \Phi_n^\top \Phi_n + \lambda I_d
\end{align}

Assume that $\forall m, \|\boldsymbol{\theta}_m\|_2 \leq S$, $\sup_{s,a} \|\boldsymbol{\phi}(s,a)\|_2 \leq Q$, $\forall m, \sup_{s,a}|\boldsymbol{\theta}_m^\top \boldsymbol{\phi}(s,a)| \leq 1$. 
Let each $U_{n,k}$ be $L$-Lipschitz in $y$: $|U(y, \boldsymbol{\phi}) - U(y', \boldsymbol{\phi})| \leq L | y - y'|$. Define
the upper confidence bound
\begin{align}
    \text{UCB}_{n}(s,a) &= 
    \sum_{k=1}^K w_{n,k} \cdot 
    U_{n,k}\Big(\hat{y}_{n, m_k}(s,a), \boldsymbol{\phi}(s,a) \Big) + \beta L \|\boldsymbol{\phi}(s,a)\|_{\Sigma_{n-1}^{-1}} \\
    \hat{y}_{n,m}(s,a) &= \boldsymbol{\phi}(s, a)^\top \hat{\boldsymbol{\theta}}_{n-1,m} \\
    \beta &= \sqrt{\lambda}S + \sqrt{2 \log\left(NM\right) + d\log\left(1 + \frac{NQ^2}{\lambda d}\right)}
\end{align}

By \cite{lattimore2020bandit} Theorem 20.5 and a union bound over $m \in [M]$, it holds that
\begin{align} 
    P\left(\exists m, \exists n \leq N :
        \|\hat{\boldsymbol{\theta}}_{n-1,m} - \boldsymbol{\theta}_m\|_{\Sigma_{n-1}} \geq \beta
    \right)
    \leq \frac{1}{N}
\end{align}
Let $E_n$ denote the event that $\forall m, \|\hat{\boldsymbol{\theta}}_{n-1,m} - \boldsymbol{\theta}_m\|_{\Sigma_{n-1}} \leq \beta$ and let $E = \bigcap_{n=1}^N E_n$. Define $H_n = \{(\boldsymbol{\phi}_{n'}, \mathbf{y}_{n'})\}_{n' = 1}^n$ to be the history up until time $n$. 
We have that
\begin{align}
    \text{BR}(N) &= 
    \mathbb E\left[ 
        \sum_{n=1}^N r_n(s_n, a_{s_n}^*) - r_n(s_n, a_n)
    \right] \\
    &= 
    \mathbb E\left[ \mathbbm{1}\{E^c\}
        \sum_{n=1}^N r_n(s_n, a_{s_n}^*) - r_n(s_n, a_n)
    \right] + 
    \mathbb E\left[ \mathbbm{1}\{E\}
        \sum_{n=1}^N r_n(s_n, a_{s_n}^*) - r_n(s_n, a_n)
    \right] \label{eq:intermed-1} \\
    &= 
    2LN \cdot \mathbb E\left[ \mathbbm{1}\{E^c\}\right] + 
    \mathbb E\left[ \mathbbm{1}\{E\}
        \sum_{n=1}^N r_n(s_n, a_{s_n}^*) - r_n(s_n, a_n)
    \right]  \\
    &\leq 2L + \mathbb E\left[ 
        \sum_{n=1}^N 
        \mathbb E \Big[
            \mathbbm{1}\{E_n\} \left(
                r_n(s_n, a_{s_n}^*) - r_n(s_n, a_n)
            \right)
        \ | \  H_{n - 1 } \Big]
    \right] \label{eq:bayes-regret-intermediate}
\end{align}
Note that the first term in Eq. \ref{eq:intermed-1} is bounded by $2LN$ since $|r_n(s,a) - r_n(s',a')| \leq L\sum_k w_{n,k} |y_{m_k}(s,a) - y_{m_k}(s',a')| \leq 2L$. \\

Define $a_{s_n}^\star = \argmax_a r_n(s_n, a)$ to be the optimal action in state $s_n$.
Noting that $P(a_{s_n}^\star = \cdot | H_n) = P(a_n = \cdot | H_n)$, we have that
\begin{align}
    &\mathbb E \left[
            \mathbbm{1}\{E_n\} \Big(
                r_n(s_n, a_{s_n}^*) - r_n(s_n, a_n)
            \Big)
        \ | \  H_{n - 1 } \right] \\
    &=\mathbb E \left[
            \mathbbm{1}\{E_n\} \Big(
                r_n(s_n, a_{s_n}^*) - \text{UCB}_{n}(s_n,a_n)
                + \text{UCB}_{n}(s_n,a_n) - r_n(s_n, a_n)
            \Big)
        \ | \  H_{n - 1 } \right]  \\
    &=\mathbb E \left[
            \mathbbm{1}\{E_n\} \Big(
                r_n(s_n, a_{s_n}^*) - \text{UCB}_{n}(s_n, a_{s_n}^*)
                + \text{UCB}_{n}(s_n,a_n) - r_n(s_n, a_n)
            \Big)
        \ | \  H_{n - 1 } \right] \\
    &\leq \mathbb E \left[
            \mathbbm{1}\{E_n\} \Big(
                \text{UCB}_{n}(s_n,a_n) - r_n(s_n, a_n)
            \Big)
        \ | \  H_{n - 1 } \right] \label{eq:ucb-ineq}
\end{align}
The inequality in Eq. \ref{eq:ucb-ineq} follows from the fact that $\text{UCB}_n(s,a) \geq r_n(s,a)$ given $E_n$. Continuing, we find that
\begin{align}
    &\mathbb E \left[
            \mathbbm{1}\{E_n\} \Big(
                \text{UCB}_{n}(s_n,a_n) - r_n(s_n, a_n)
            \Big)
        \ | \  H_{n - 1 } \right]  \\
    &\leq \mathbb E \left[
            \mathbbm{1}\{E_n\} 
            \left(
            \sum_{k=1}^K w_{n,k} \left(
            U_{n,k}(\hat{y}_{n,m_k}, \boldsymbol{\phi}_n) - 
            U_{n,k}(y_{n,m_k}, \boldsymbol{\phi}_n)
            \right)
            + \beta L \|\boldsymbol{\phi}_n\|_{\Sigma_{n-1}^{-1}}
            \right)
        \ | \  H_{n - 1 } \right] \\
    &\leq \mathbb E \left[
            \mathbbm{1}\{E_n\} 
            \left(
            \sum_{k=1}^K w_{n,k} L |\hat{y}_{n,m_k} -
            y_{n,m_k}|
            + \beta L \|\boldsymbol{\phi}_n\|_{\Sigma_{n-1}^{-1}}
            \right)
        \ | \  H_{n - 1 } \right] \\
    &= \mathbb E \left[
            \mathbbm{1}\{E_n\} 
            \left(
            \sum_{k=1}^K w_{n,k} L |
            \langle \boldsymbol{\phi}_n, \hat{\boldsymbol{\theta}}_{n-1, m_k} - \boldsymbol{\theta}_{m_k} \rangle
            |
            + \beta L \|\boldsymbol{\phi}_n\|_{\Sigma_{n-1}^{-1}}
            \right)
        \ | \  H_{n - 1 } \right] \\
    &\leq \mathbb E \left[
            \mathbbm{1}\{E_n\} 
            \left(
            \sum_{k=1}^K w_{n,k} L 
            \|\boldsymbol{\phi}_n\|_{\Sigma_{n-1}^{-1}} 
            \|\hat{\boldsymbol{\theta}}_{n-1, m_k} - \boldsymbol{\theta}_{m_k}\|_{\Sigma_{n-1}} 
            + \beta L \|\boldsymbol{\phi}_n\|_{\Sigma_{n-1}^{-1}}
            \right)
        \ | \  H_{n - 1 } \right] \\
    &\leq \mathbb E \left[
            2L\beta  \|\boldsymbol{\phi}_n\|_{\Sigma_{n-1}^{-1}}
        \ | \  H_{n - 1 } \right]
\end{align}
Plugging the above into Eq. \ref{eq:bayes-regret-intermediate} combined with $\mathbb E \left[ r_n(s_n, a_{s_n}^*) - r_n(s_n, a_n) \right] \leq 2L$, we conclude that
\begin{align}
    \text{BR}(N) &\leq 2L + 
    2L\beta  \cdot \mathbb E \left[
        \sum_{n=1}^N
         \min\left(1, \|\boldsymbol{\phi}_n\|_{\Sigma_{n-1}^{-1}}\right)
    \right] \\
    &\leq 2L + 
    2L\beta \cdot \sqrt{
    N \cdot \mathbb E \left[
        \sum_{n=1}^N
         \min\left(1, \|\boldsymbol{\phi}_n\|_{\Sigma_{n-1}^{-1}}^2\right)
    \right]} \label{eq:cauchy} \\
    &\leq 2L + 
    2L\beta \cdot \sqrt{
    2Nd 
        \log\left(1 + \frac{NQ^2}{\lambda d}\right)
    } \label{eq:ellip-potential} \\
    &= 2L \left( 
    1 + 
    \left(\sqrt{\lambda}S + \sqrt{2 \log\left(NM\right) + d\log\left(1 + \frac{NQ^2}{\lambda d}\right)}\right)
    \sqrt{
    2Nd 
        \log\left(1 + \frac{NQ^2}{\lambda d}\right)
    }
    \right)
\end{align}
Eq. \ref{eq:cauchy} follows from Cauchy-Schwarz and Eq. \ref{eq:ellip-potential} follows from the elliptical potential lemma (\cite{lattimore2020bandit} Lemma 19.4).\\

In conclusion, under the assumption that $Q, S, \lambda$ are $O(1)$, we have that the Bayesian regret is of order $O(Ld \sqrt{N \log(NM)\log(N/d)})$

\section{Step Count Simulator}
\label{appendix:heartsteps}
Here we discuss the details of the simulator. We model one outcome $(M=1)$, step count, and assume the step count outcome follows a linear model,
\begin{equation}
    y_{i, t} = \boldsymbol{\phi}_{i,t}^\top \boldsymbol{\theta}^\star + \varepsilon
\end{equation}
where $\varepsilon \sim \mathcal{N}(0,1)$.
The featurization $\boldsymbol{\phi}_{i,t}=[1, y_{i, t-1}, t_a, t_i, nn]$ contains two components that represent the treatment effect for each of the two groups ($t_a$ for active users and $t_i$ for inactive users) and one counter that accumulates the number of notifications ($nn$). The treatment effect is represented as a decaying sigmoid of the number of notifications. For the active group it is
\begin{equation}
   t_a(nn) = 1 - \sigma\left(
        \left(nn - \frac{T}{0.95}\right) \cdot \frac{5}{T}
    \right) 
\end{equation}
and for the inactive group it is
\begin{equation}
   t_i(nn) = 1 - \sigma\left(
        \left(nn - \frac{T}{0.65}\right) \cdot \frac{2}{T}
    \right) 
\end{equation}
We set $\boldsymbol{\theta}^\star = [-0.04, 0.9999, 0.3, 0.15, 0]$, which ignores the number of notifications that have been sent to a user. 
\begin{figure}[htbp]
    \centering
    \includegraphics[width=0.85\linewidth]{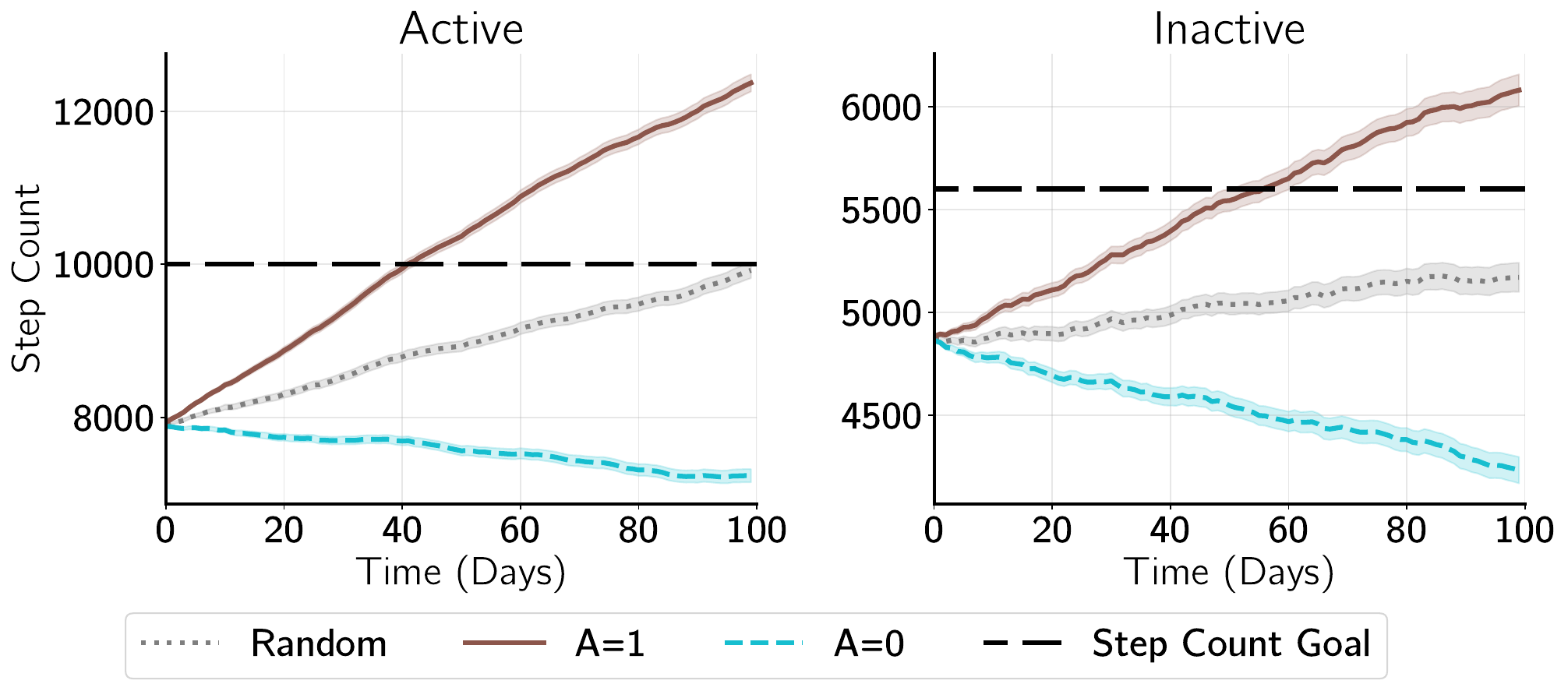}
    \caption{We compare step count dynamics between active (left) and inactive (right) groups. We plot the simulated step count for each group under three policies: a policy that always sends a notification $(A=1)$, a policy that never sends notifications $(A=0)$, and a random policy. Each groups step count goal is plotted as a dashed horizontal line.}
    \label{fig:te_groups}
\end{figure}

The first utility function upweights step count and is given by
\begin{equation}
\label{eq:u1}
U_{i,t,1}(y_{i,t}, \boldsymbol{\phi}_{i,t})=
 \begin{cases} 
      \alpha_1 \cdot y_{i,t} & y_{i,t}\leq \text{step goal} \\
      \alpha_2 \cdot y_{i,t} + \gamma & y_{i,t} > \text{step goal}
   \end{cases}
\end{equation}
where $\alpha_1=0.005, \alpha_2=0.001$, $\gamma=0.42$ if the user is an active user, and $\gamma=0.3$ if the user is an inactive user. The step count goal for active users is 10,000 steps, and for inactive users it is 5,600 steps. This utility is a piece-wise linear function with a high slope before a user achieves their step count goal, and a much smaller slope afterwards.
The second utility function is given by
\begin{equation}
    U_{i,t,2}(y_{i,t}, \boldsymbol{\phi}_{i,t})=-\beta((nn)^2)
\end{equation}
where $\beta = 0.00003$. This utility decreases quadratically as the number of notifications increases. 

\section{Gym Attendance Semi-synthetic simulator}
\label{appendix:24h}
\subsection{Gym Attendance Regression Model}
\label{appendix:24h:simulator}
The dataset from \cite{milkman2021megastudies}\footnote{Available for download at \url{https://osf.io/9av87/?view_only=8bb9282111c24f81a19c2237e7d7eba3}} contains weekly gym visits for 61,293 participants. For each participant, the data contains demographic information (age, gender, new member status, state of residence) along with weekly gym visits for each of the four weeks of the main study, ten weeks-post study, and up to one year of pre-study historical data. The dataset contains 3,010,659 total entries, with an average of 49.12 weeks of data per participant. Participants were assigned to one of 54 different conditions using a cohort-based randomization scheme, including 53 interventions and one control. \\

To construct a simulator for a bandit environment, we featurize each week of data for each participant using their demographics and history. We then train a regressor to predict that week's number of gym visits given the featurization. In particular, let $y_{i,t}$ denote the number of times participant $i$ attended the gym in week $t$, let $a_{i,t}$ be the intervention condition assigned to participant $i$ at week $t$ (with $a=0$ denoting no intervention/control), and let $H_{i,t-1} = \{(a_{i,t'}, y_{i,t'})\}_{t' < t}$ denote their history up until time $t$. Let $s_{i,t}$ represent participant $i$'s state at time $t$, including their demographic information and history. We use the following featurization function:
\begin{align*}
    \phi(s_{i,t}, a_{i,t}) = \Bigg[
        & \text{age, state of residence, gender, new member status}, \\
        & \quad \text{mean}(y \in H_{i,t-1}), \text{min}(y \in H_{i,t-1}), \text{max}(y \in H_{i,t-1}), \\
        & \quad y_{i,t-1}, y_{i,t-2}, \text{longest streak, \# streaks}, \mathbbm{1}\{a_{i,t} \neq 0\}, \sum_{a \in H_{i,t}}\mathbbm{1}\{a \neq 0\}, \text{one-hot}(a_{i,t})
    \Bigg]
\end{align*}
State of residence is mapped to an index in $[0,9]$, where 0-8 correspond to the 9 most common states (CA, TX, CO, WA, OR, FL, NY, HI, NJ), covering 96\% of the dataset, and 9 for all other states. We use this integer to create a one-hot encoding of state in $\mathbb{R}^d$. New members are participants who joined 24 Hour Fitness less than one year prior to the study period. A streak is a period of two or more consecutive weeks where the participant had more than one gym visit in each week. The one-hot encoding sets the control condition to the 0 vector and all other actions to a unit vector in $\mathbb{R}^{A-1}$.\\

We reduce the action set from the original 54 actions down to six (including the control condition). To do so, we categorized the original 53 interventions into five categories (financial incentives, messages that affirm your values, notifications to plan workouts, notifications to reflect on your number of gym visits per week) and restrict ourselves to participants who were assigned the intervention with the highest treatment effect within each category (``Bonus for Returning after Missed Workouts'', ``Exercise Social Norms Shared (High and Increasing)'', ``Planning Fallacy Described and Planning Revision Encouraged'', ``Exercise Commitment Contract Explained'', ``Fitness Questionnaire with Decision Support \& Cognitive Reappraisal Prompt''; respectively). This reduces the size of our dataset to 487,856 weekly gym visits for 9,868 participants. 
Reducing the action set greatly reduces the dimensionality of our featurization (via the one-hot encoding) and accelerates bandit learning over the short, four-week study duration. It was also infeasible for us to elicit preferences over 53 different interventions in our online study (discussed below in \ref{appendix:24h:online-study}). \\ 

In addition, we exclude participants with missing data or who have less than 5 weeks of pre-study history. We split this filtered dataset into roughly equally sized train and test set by cohort, using the first 13 study cohorts (ordered by date of study enrolled) for training (271,948 samples) and the last 14 cohorts for testing (184,979 samples). \\

In Table \ref{tab:regressors_mse} below, we report the root mean squared error of various regressors (implemented using scikit-learn\footnote{\url{https://scikit-learn.org}} using the default hyperparameters) on the test set of the original action set and on the reduced action set. We use an identical featurization, noting that the dimensionality is larger when using the original action set ($D=75$) than the reduced action set ($D=27$) due to the one-hot action featurization. We also note that the train (1,715,186) and test (1,096,798) sets are larger when using the original action set. 

\begin{table}[ht]
\centering
\begin{tabular}{lcc}
\toprule
\textbf{Regressor}            & \textbf{Reduced Action Set ($A=6$)} & \textbf{Original Action Set ($A=54$)} \\
\midrule
Linear Regression    & 1.05 & 1.06 \\
Random Forest        & 1.10 & 1.10 \\
Gradient Boosting    & 1.04 & 1.05 \\
\bottomrule
\end{tabular}
\caption{Root mean squared error (RMSE) on the test set for different regressors.}
\label{tab:regressors_mse}
\end{table}

We find that reducing the action set leads to a very small reduction in RMSE, indicating that reducing the action set does not lead to substantial change in the predictive accuracy of our models. We also find that a gradient boosting regressor has the best performance, very closely followed by linear regression. We use a gradient boosting regressor in our simulator due to its slight performance advantage, but use a linear model in our bandit algorithms as it admits a closed-form posterior. \\

\begin{figure}[h]
    \centering
    \includegraphics[width=0.6\textwidth]{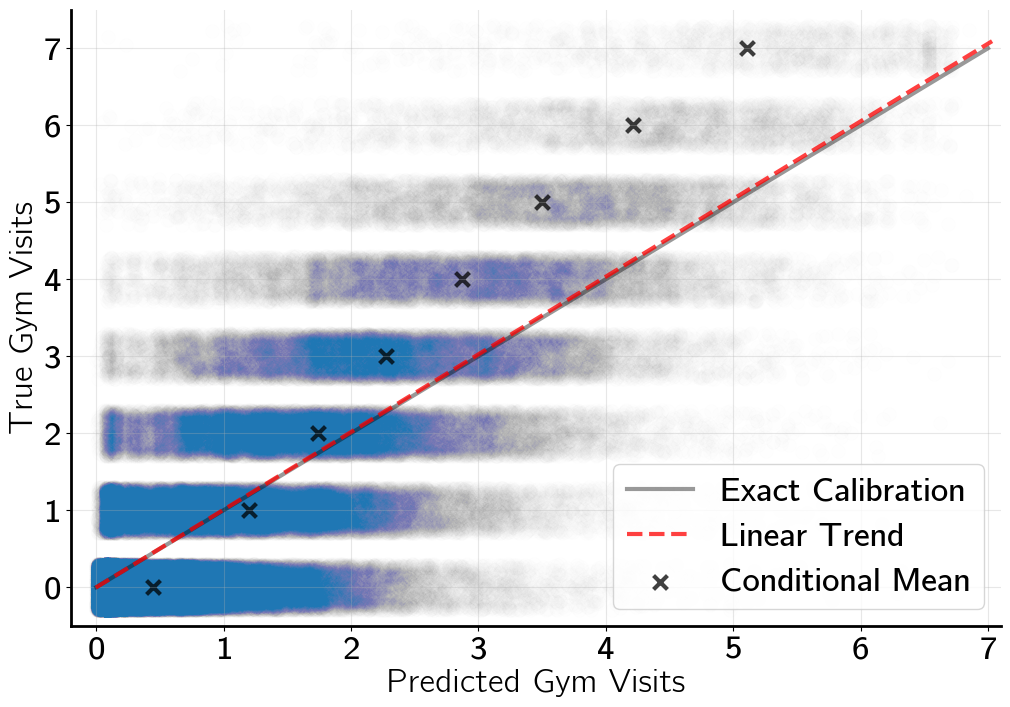}
    \caption{Calibration plot for the gradient boosting regressor on the test set.}
    \label{fig:24h-calibration}
\end{figure}

In Figure \ref{fig:24h-calibration}, we show a calibration plot of our gradient boosting regression model's predicted number of gym visits compared to the true number of gym visits in the test set. 
While a linear trend line suggests near perfect calibration, examining conditional means (i.e., $\mathbb{E}[\hat{y}_{i,t} | y_{i,t} = k]$) our model has positive bias when the true number of visits is less than 2, and negative bias when the true number of visits is greater than or equal to 2.
This is likely due to the heavily skewed distribution of gym visits in the dataset (1.28 mean, 0.0 median).

Since our regression model produces continuous outcomes, while gym visits are integers, we convert a scalar model output $\hat{y}_{i,t}$ to an integer-valued $y_{i,t}$ using the following formula.
\begin{equation*}
    y_{i,t} \sim \hat{y}_{i,t} - \lfloor y_{i,t} \rfloor + \text{Bernoulli}\left(1 - \left(\hat{y}_{i,t} - \lfloor y_{i,t} \rfloor \right)\right)
\end{equation*}
This conversion does not affect our experimental results other than adding noise.

\subsection{Online Preference Survey}
\label{appendix:24h:online-study}
To gather preferences about the different interventions, we conducted an online study on Prolific. Our survey was delivered via Qualtrics and took about 6 minutes to complete. Participants were compensated \$2 for completing the survey. Our survey and recruitment procedures were approved by our university's institutional review board. \\

Our screening criteria required participants to be fluent in English, have an approval rating of 95 or above, and have at least 50 previous submissions. We conducted one pilot survey with 20 participants and a full survey with 200 participants one month later. The pilot survey did not include questions about $\beta$ but was otherwise identical up to small changes in wording. We exclude participants who failed attention checks and participants who completed the survey less than one standard deviation below the mean completion time (less than 150 seconds). We also exclude one participant who had no match in test dataset (who reported going to the gym seven days per week), leaving 209 final participants. We report participant demographic information in Table \ref{tab:demographics} and plot the geographic distribution in Figure \ref{fig:map} below.\\

\begin{table}[ht]
\centering
\begin{tabular}{ll}
\toprule
\textbf{Age} & 
    \textit{Mean:}  36.93, 
    \textit{Median:} 33, 
    \textit{SD:} 13.03, 
    \textit{Min:} 18, 
    \textit{Max:} 75 \\
\textbf{Gender} & 
    \textit{Male:} 100,
    \textit{Female:} 95,
    \textit{Non-binary/Genderqueer/Non-conforming:} 12, \\
    & \textit{Transgender male:} 1,
    \textit{Transgender female:} 1
\\
\textbf{Ethnicity} & 
    \textit{White:} 154,
    \textit{Black:} 18,
    \textit{Asian:} 16,
    \textit{Mixed:} 10,
    \textit{Other:} 8
\\
\bottomrule
\end{tabular}
\caption{Demographic information for the $N=209$ participants in our online survey.}
\label{tab:demographics}
\end{table}

\begin{figure}[h]
    \centering
    \includegraphics[width=0.7\textwidth]{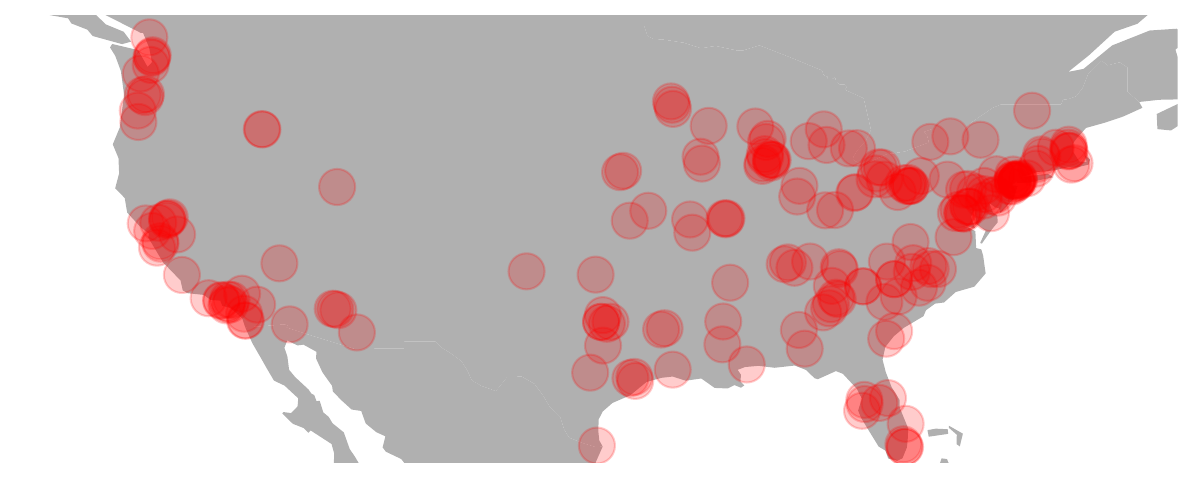}
    \caption{Geographic distribution of the $N=209$ participants in our online survey.}
    \label{fig:map}
\end{figure}

The survey first asked participants whether they currently go to the gym and branched into two analogous conditions based on their responses. For participants that go to the gym, the survey asked participants how many days per week they currently go to the gym and how often they would like to go. It then asked participants, \textit{``For each of the following, indicate the degree to which you think the following reminders or incentives would motivate you to increase your gym attendance''} for each of the five aforementioned intervention categories on a 5-point Likert scale (1: Not at all increase, 5: Extremely increase). For participants who did not go to the gym, the survey asked how many days per week they engage in physical activity, how many days they would like to engage in physical activity, and analogous preference questions indicating \textit{``the degree to which you think the following reminders or incentives would motivate you to increase your levels of physical activity''}. We then asked \textit{``If an artificial intelligence (AI) system were to make recommendations for increasing my [gym attendance/levels of physical activity], I would listen to the AI’s advice.''} \\

To extract preference scores $\alpha$ from the Likert scale responses, we mapped each response to an integer (0: Not at all increase, 4: Extremely increase) and set the control condition to 0. $\alpha$ was computed using a softmax over this preference vector. To compute $\beta$, we again mapped responses to an integer $b$ (0: Strongly agree, 4: Strongly disagree) and transformed this to the $[0.1, 0.9]$ range using $\beta = 0.1 + (0.8 \cdot b)/4$. \\

The last block of the survey included general questions about how many minutes per week participants engage in physical activity, the types of activities they engage in, and the times of day they engage in physical activity. We also asked participants free-responses questions about their short-term and long-term physical activity goals. A full analysis of the survey responses is beyond the scope of this work, though relevant findings are highlighted in the main text.\\

To create our bandit simulator, we match each participant $i$ in our online study to a participant $j$ in the gym attendance dataset. First, we filter participants in the test set of our gym attendance dataset based on gender, age (within 5 years), new member status, and their average number of gym visits pre-intervention (within 0.5). We treat participants in the online study who do not go to the gym as new members and set their pre-intervention attendance to 0. We then randomly select among the participants in the test set that meet the matching criteria. For participants in the pilot study who did not have a response to this question, we randomly sample a $\beta$ from participants in the full study.
A new random match is generated each simulation to demonstrate that our algorithm can robustly optimize over any choice of preferences drawn from a realistic distribution.

\end{document}